\pdfoutput=1
\documentclass[11pt]{article}

\usepackage[preprint]{acl}

\usepackage{times}
\usepackage{latexsym}

\usepackage[T1]{fontenc}
\usepackage[utf8]{inputenc}

\usepackage{microtype}

\usepackage{inconsolata}
\usepackage{amsmath}
\usepackage{xcolor}
\usepackage{microtype}
\usepackage{hyperref}
\usepackage{cleveref}
\usepackage{url}
\usepackage{booktabs}
\usepackage{amsfonts}
\usepackage{pifont}
\usepackage[table]{xcolor}
\usepackage{longtable}
\usepackage{graphicx}
\usepackage{xspace}
\usepackage{tabularx}
\usepackage{multirow}
\usepackage{tcolorbox}
\tcbuselibrary{breakable}
\usepackage{caption}
\usepackage[normalem]{ulem}
\usepackage{arydshln}
\usepackage{titletoc}
\usepackage{lipsum}
\usepackage[shortlabels]{enumitem}
\usepackage{array}
\usepackage{kotex}
\usepackage{graphicx}

\newcommand{\datasetname}{Everyday Physics in Korean Contexts\xspace}
\newcommand{\dn}{EPiK\xspace}

\useunder{\uline}{\ul}{}

\crefformat{section}{\S#2#1#3}
\crefformat{subsection}{\S#2#1#3}
\crefformat{subsubsection}{\S#2#1#3}
\crefrangeformat{section}{\S\S#3#1#4 to~#5#2#6}
\crefmultiformat{section}{\S\S#2#1#3}{ and~#2#1#3}{, #2#1#3}{ and~#2#1#3}

\title{Everyday Physics in Korean Contexts: A Culturally Grounded Physical Reasoning Benchmark}

\author{
    Jihae Jeong$^{1,}$\Thanks{Both authors contributed equally to this work.}
    \quad 
    DaeYeop Lee$^{1,2,*}$ 
    \quad  
    DongGeon Lee$^{1}$ 
    \quad  
    Hwanjo Yu$^{1,}$\thanks{Corresponding author.} \\
        $^1$POSTECH \quad
        $^2$MODULABS \\
    \texttt{\{wisdomjeong, daeyeoplee, donggeonlee, hwanjoyu\}@postech.ac.kr}
}

\begin{document}
\maketitle

\begin{abstract}

Existing physical commonsense reasoning benchmarks predominantly focus on Western contexts, overlooking cultural variations in physical problem-solving. To address this gap, we introduce \dn (\datasetname), a novel benchmark comprising 181 binary-choice problems that test physical reasoning within Korean cultural contexts, ranging from kimchi (Korean food) to traditional fermentation.
\dn is constructed using a two-stage generation and verification pipeline to create culturally-authentic problems across 9 reasoning subtasks and 84 scenarios. Unlike approaches based on simple translation, our method generates problems organically from Korean contexts while upholding rigorous physical reasoning standards.
Our evaluations show that Korean-specialized models consistently outperform general-purpose models of comparable size.
This performance gap highlights the limitations of culturally-agnostic models and demonstrates the critical need for culturally-aware benchmarks to truly measure language understanding. 
Our \dn is publicly available at \url{https://huggingface.co/datasets/jjae/EPiK}.

\end{abstract}
\section{Introduction}

Large Language Models (LLMs) have recently shown impressive capabilities in reasoning about the physical world, yet their performance remains uneven across different contexts \cite{Ismayilzada-2023-CRoW, Wang-2023-NEWTON}.
Physical commonsense reasoning, the ability to understand and predict everyday physical phenomena, is fundamental to human intelligence and a critical capability for AI systems that interact with the real world \cite{Pensa-2024-Multi}.
While recent benchmarks \cite{Bisk-2020-PIQA, Wang-2023-NEWTON} have achieved substantial progress in evaluating this capability, they predominantly focus on Western, English-speaking contexts, leaving a significant gap in our understanding of how AI systems perform across diverse cultural and linguistic environments \cite{Ponti-2020-XCOPA, Shi-2024-CultureBank}.

This gap is particularly problematic because physical reasoning, despite appearing universal, is deeply intertwined with cultural context \cite{Acharya-2020-Towards, Shi-2024-CultureBank, Acquaye-2024-Susu, Shen-2024-Understanding}. 
Consider the task of `` keeping the house warm during winter.''
While Western-centric benchmarks might emphasize fireplaces, Korean commonsense includes understanding combustion dynamics in \textit{ondol} (온돌, Korean traditional heating system) and the role of \textit{a-gung-i} (아궁이, Korean traditional fireplace). 
Such culturally-embedded physical knowledge \cite{Acharya-2020-Towards, Liu-2025-Culturally} is essential for AI systems deployed in Korean contexts, yet remains unmeasured by existing benchmarks.

Current evaluation practices for Korean language models typically rely on translated versions of English benchmarks, which introduce multiple limitations \cite{Sakai-2024-mCSQA}. 
First, translation often distorts the physical scenarios being described, as certain concepts lack direct equivalents across languages \cite{Artetxe-2020-Translation, Artetxe-2023-RevisitingMT}. 
Second, translated benchmarks fail to capture Korea-specific physical phenomena that arise from unique environmental conditions (e.g., distinct monsoon patterns), living arrangements (e.g., floor heating systems), and material culture (e.g., traditional cooking implements). 
Third, this approach perpetuates a Western perspective of \textit{universal} physical reasoning, potentially overlooking diverse problem-solving strategies that emerge from different cultural contexts \cite{Hershcovich-2022-Challenges, Koto-2024-IndoCulture, Myung-2024-BLEnD, Liu-2025-Culturally}.

\begin{figure*}[ht!]
    \centering
    \includegraphics[width=\linewidth]{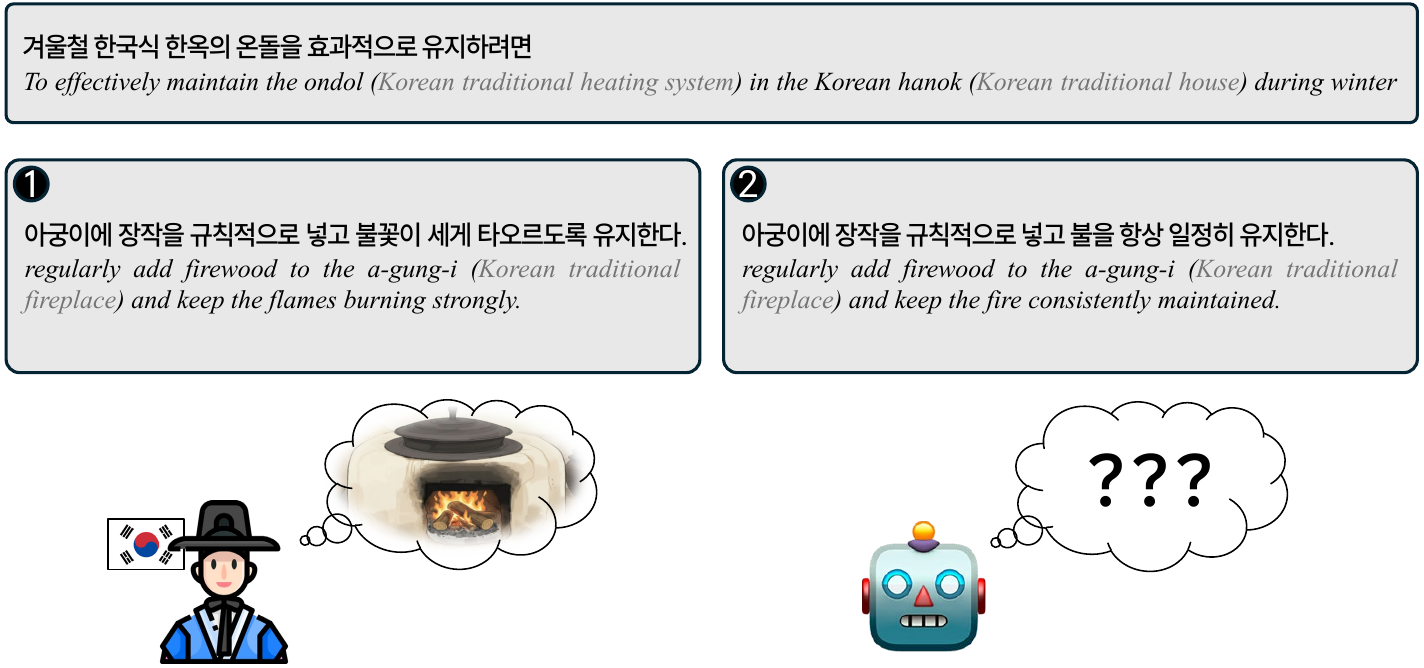}
    \caption{An illustrative example from our proposed \dn benchmark. The problem requires understanding the traditional Korean \textit{ondol} heating system, demonstrating the need to integrate physical reasoning with specific cultural contexts, a task that can be especially challenging for multi-lingual language models.}
    \label{fig:dataset_example}
\end{figure*}

To address these limitations, we introduce the \dn (\datasetname) benchmark, a comprehensive dataset designed to evaluate physical commonsense reasoning within authentic Korean contexts. 
\dn consists of 181 carefully curated binary-choice problems that require an understanding of both universal physical principles and Korea-specific applications. 
Unlike simple translations, our questions are generated from the ground up to reflect genuine Korean daily life scenarios while maintaining rigorous standards for physical reasoning complexity.

Our contributions are summarized as follows:

\vspace{-0.075in}

\begin{itemize}[itemsep=0.3mm, parsep=1pt, leftmargin=*]
    \item We propose a rigorous methodology for building a culturally-grounded benchmark, featuring a systematic taxonomy that bridges universal physical principles with Korean-specific contexts and a two-stage pipeline with interactive verification and bias filtering to ensure question validity.
    \item Through extensive experiments on a variety of models, we demonstrate a significant performance gap between general-purpose and Korean-specialized models, highlighting the critical need for culturally-aware evaluation.
    \item We publicly release \dn as a valuable dataset to facilitate research on culturally-aware reasoning and to promote the development of more inclusive and globally competent models.
\end{itemize}

Our results demonstrate that Korean-specialized models consistently outperform general models on tasks requiring Korean-contextualized physical reasoning.
This gap underscores the need for evaluation frameworks that respect linguistic and cultural diversity in assessing AI capabilities.
Beyond Korean contexts, \dn serves as a framework for developing culturally-grounded benchmarks in other languages, contributing to a more comprehensive understanding of physical commonsense reasoning across human cultures.

\section{Related Work}

\subsection{Physical Commonsense Reasoning Benchmark}
Physical commonsense reasoning has emerged as a critical challenge in natural language understanding (NLU) \cite{Davis-2024-Survey}.
PIQA \cite{Bisk-2020-PIQA} introduced physical commonsense as a textual reasoning task where, given a goal from everyday life, one must choose the more plausible method among two candidates.
Despite being easy for humans, PIQA exposed a large gap for pretrained models, attributing difficulty to reporting bias in text-only pretraining \cite{Paik-2021-Octopus}. 

Our benchmark follows the PIQA two-choice format and the same core objective (plausible physical action selection) but grounds both questions and answers in Korean daily life, thus serving as a `Korean Cultural PIQA'.

\begin{figure*}[ht!]
    \centering
    \includegraphics[width=\linewidth]{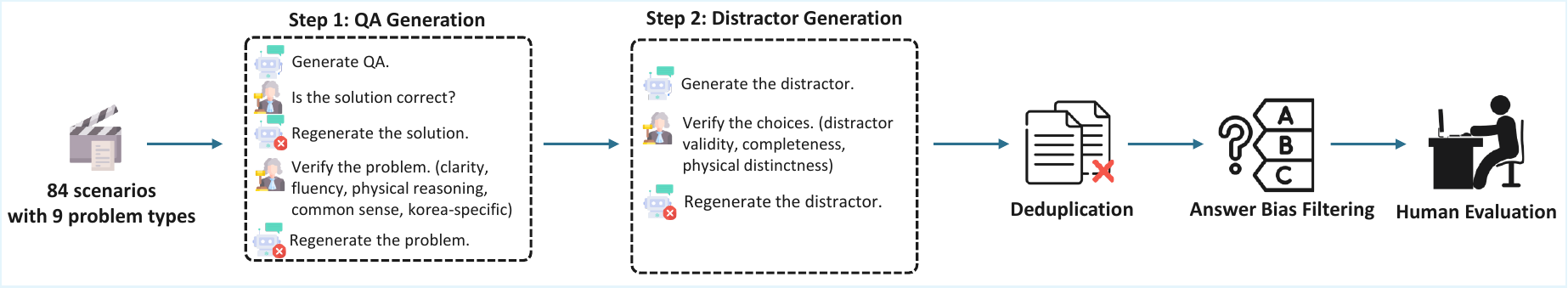}
    \vspace{-5mm}
    \caption{An overview of \dn construction process. 
    \textbf{1) Foundational Framework}: Define 84 scenarios and 9 problem types. 
    \textbf{2) Two-stage Problem Generation and Verification}: Generate QA samples and verify the problems. 
    \textbf{3) Deduplication} 
    \textbf{4) Answer Bias Filtering} 
    \textbf{5) Human Evaluation} }
    \label{fig:dataset_construction}
\end{figure*}

\subsection{Korean Commonsense Benchmark}
Most commonsense evaluations have been English-centric or translation-driven \cite{Ponti-2020-XCOPA, Lin-2020-Common}. 
Multilingual commonsense benchmarks often arise via translation or projection from English resources, which can import artifacts and dilute culturally prototypical solutions \cite{Nie-2024-Multilingual}.

Recent efforts emphasize culturally grounded assessments, particularly for Korean language resources. 
CLIcK~\cite{Kim-2024-CLIcK} collects language-and-culture QA from official exams and textbooks. 
KoCommonGEN v2\cite{Seo-2024-KoCommonGENv2} takes a different approach by reconstructing a commonsense generation dataset from scratch. 

Complementary to these, \dn is built from the ground up on Korean daily-life physical reasoning rather than translation, targeting safety-, tool-, and procedure-sensitive decisions that users routinely face.
While Ko-PIQA \cite{choi2025kopiqa} also focuses on physical commonsense in a cultural context, our work offers two key distinctions. First, our dataset is fully model-generated, providing a scalable and efficient approach to data creation. Second, we employ a rigorous ablation-based filtering method to remove instances solvable by superficial cues, thereby ensuring that our benchmark truly requires genuine reasoning and is free from statistical artifacts.
\section{Dataset Construction}

\dn is designed to evaluate physical commonsense reasoning within the unique context of Korean culture. 
The problems are formulated to be solvable by native Korean speakers with general everyday knowledge of traditional and contemporary life.
The dataset follows a two-alternative multiple-choice format, requiring a model to select the correct solution from two candidates, following the style of PIQA.
The entire construction pipeline is illustrated in Figure~\ref{fig:dataset_construction}.

\subsection{Foundational Framework: Scenarios and Task Taxonomy}
To ensure comprehensive coverage and diversity, we first establish a foundational framework consisting of background scenarios and a taxonomy of physical reasoning tasks.

\paragraph{Task Taxonomy}
To systematically guide the problem generation process, we establish a task taxonomy inspired by the PIQA dataset \cite{Bisk-2020-PIQA}. 
It comprises four high-level categories and nine fine-grained sub-categories capturing a wide range of physical reasoning skills. This taxonomy, detailed in Table~\ref{tab:problem_taxonomy}, serves as a structured framework to ensure a balanced distribution of reasoning types. However, it is important to note that the generated questions often integrate concepts from multiple categories.
\begin{table*}[h!]
    \centering
    \small

    \resizebox{\textwidth}{!}{%
        \begin{tabular}{l@{\hspace{0.85\tabcolsep}}p{13.5cm}}
        \toprule
        \textbf{Problem Type}                     & \textbf{Definition}                                                                                        \\ \midrule
        \multicolumn{2}{c}{\cellcolor{gray!25}\textit{\textbf{기초 지식 및 속성 이해 (Basic Knowledge and Attribute Comprehension)}}}                                                      \\ \midrule
        물리 개념 및 원리                                & 기본적인 물리 법칙과 원리에 대한 이해를 평가합니다.                                                                              \\
        \textit{Physical Concepts and Principles} & \textit{Assesses understanding of fundamental physical laws and principles.}                               \\ \midrule
        객체 속성 및 기능                                & 사물의 재질, 무게, 형태, 질감 등 고유 속성과 가능한 상호작용을 이해하고 평가합니다.                                                          \\
        \textit{Object Attributes and Functions} &
          \textit{Assesses understanding and evaluation of inherent properties like material, weight, shape, and texture, and potential interactions.} \\ \midrule
        물질의 상태와 변화                                & 고체, 액체, 기체 상태의 특징과 녹는 것, 어는 것, 끓는 것, 녹이는 것 등 물질의 상태 변화의 원리를 이해하고 평가합니다.                                    \\
        \textit{States of Matter and Changes} &
          \textit{Assesses understanding and evaluation of the characteristics of solid, liquid, and gas states, and the principles of state changes like melting, freezing, boiling, and dissolving.} \\ \midrule
        \multicolumn{2}{c}{\cellcolor{gray!25}\textit{\textbf{인과관계 및 동적 추론 (Causality and Dynamic Reasoning)}}}                                                                   \\ \midrule
        물리적 결과 예측                                 & 특정 행동이나 사건이 가져올 물리적 결과를 예측하는 능력을 평가합니다.                                                                    \\
        \textit{Predicting Physical Outcomes}     & \textit{Assesses the ability to predict the physical outcomes of specific actions or events.}              \\ \midrule
        물리적 원인 분석                                 & 이미 발생한 현상의 원인을 가장 타당한 물리적 관점에서 설명하는 능력을 평가합니다.                                                             \\
        \textit{Analyzing Physical Causes} &
          \textit{Assesses the ability to explain the cause of an already-occurred phenomenon from the most plausible physical perspective.} \\ \midrule
        \multicolumn{2}{c}{\cellcolor{gray!25}\textit{\textbf{목표 지향 및 응용 추론 (Goal-Oriented and Applied Reasoning)}}}                                                              \\ \midrule
        도구 및 절차 활용                                & 목표 달성을 위해 적절한 도구를 선택하고 올바른 절차에 따라 사용하는 능력을 평가합니다.                                                          \\
        \textit{Utilizing Tools and Procedures}   & \textit{Assesses the ability to select appropriate tools and follow correct procedures to achieve a goal.} \\ \midrule
        문제 해결 및 계획                                & 주어진 문제를 해결하기 위해 여러 단계의 물리적 행동을 계획하고 실행하는 능력을 평가합니다.                                                        \\
        \textit{Problem Solving and Planning}     & \textit{Assesses the ability to plan and execute multi-step physical actions to solve a given problem.}    \\ \midrule
        물리적 타당성 판단                                & 제시된 해결책이 물리적으로 가능한지, 현실 세계의 상식적인 도구와 개념에 기반하는지를 평가합니다.                                                     \\
        \textit{Judging Physical Plausibility} &
          \textit{Assesses whether a proposed solution is physically possible and based on common-sense tools and concepts in the real world.} \\ \midrule
        \multicolumn{2}{c}{\cellcolor{gray!25}\textit{\textbf{상황 및 제약 조건 추론 (Situational and Constraint Reasoning)}}}                                                             \\ \midrule
        위험성 및 안전성 평가                              & 잠재적 위험을 예측하고 안전을 확보하기 위한 방법을 추론하는 능력을 평가합니다.                                                               \\
        \textit{Risk and Safety Assessment}       & \textit{Assesses the ability to predict potential risks and infer methods to ensure safety.}               \\ \bottomrule
        \end{tabular}%
        }
    \caption{The detailed taxonomy of the nine problem types designed for the \dn. Our classification scheme is organized into four coarse-grained reasoning categories, each containing several fine-grained sub-categories.}
    \label{tab:problem_taxonomy}
\end{table*}

\begin{figure}[ht!]
    \centering
    \includegraphics[width=\linewidth]{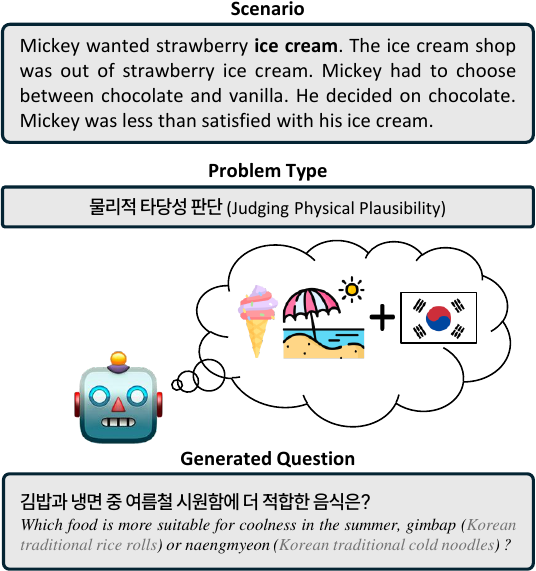}
    \caption{Illustration of the question generation process based on scenario and problem type.}
    \label{fig:scenario_example}
\end{figure}

\paragraph{Scenario Sourcing}
To anchor our questions in diverse and realistic contexts, we source background scenarios from the TRACIE dataset \cite{Zhou-2021-Temporal}.
We extract and deduplicate the premises from its training set, yielding a pool of 84 unique scenarios.
These temporally rich situations serve as scaffolds for physically grounded reasoning.
For each scenario, we systematically generate questions corresponding to our nine task types, ensuring broad coverage of both situational contexts and reasoning challenges.

\subsection{Two-Stage Generation and Verification Pipeline}
We implement a two-stage pipeline to generate high-quality questions with unambiguous answers and plausible distractors.
This approach decouples the creation of the correct solution from the distractor, allowing for rigorous quality control at each step. Using this pipeline, we generated a total of 558 samples. Details are in Appendix~\ref{appendix: Dataset Construction Details}, and the full prompts can be found in Appendix~\ref{appendix:Prompt Details}.

\subsubsection{Stage 1: Question and Correct Answer Generation}
\paragraph{Generation} 
The generation process is guided by a background scenario and a target question type from our taxonomy, with an example illustrated in Figure~\ref{fig:scenario_example}. The goal is to produce question-answer pairs where the solution requires an understanding of Korean cultural context for physical reasoning. To ensure stylistic diversity, we randomized parameters such as the maximum length of both the question and the answer (ranging from 10 to 24 characters, with a minimum combined length of 25 characters), and the final punctuation style of the question.

\paragraph{Verification}
Each generated question-answer pair underwent a rigorous two-part verification process:

\paragraph{1) Answer-Centric Verification}
This stage validates the correctness of the answer. If an answer is incorrect, the verifier provides a detailed rationale. This rationale serves as a corrective signal for the generator, which then produces a revised answer, creating an effective feedback loop.

\paragraph{2) Question-Centric Verification} 
The stage assesses the quality of the question itself based on five criteria: (a) clarity, (b) presence of a physical reasoning component, (c) solvability with general Korean commonsense, (d) integration of Korean cultural elements, and (e) linguistic fluency. A question-answer pair was discarded if it failed any criterion after three regeneration attempts.

\subsubsection{Stage 2: Distractor Generation}
\paragraph{Generation} 
To create a compelling distractor, the generator introduces minimal but critical changes to the correct answer, applying slight lexical or phrasal perturbations while preserving sentence structure and style. This approach produces challenging negative examples that require a nuanced understanding of the underlying physical principle. 

\paragraph{Verification} 
Each generated distractor undergoes a dedicated verification procedure to ensure it meets a set of strict standards. 
The criteria for a valid distractor are as follows:

\vspace{-0.075in}

\begin{enumerate}[itemsep=0.3mm, parsep=1pt, leftmargin=*]
\item It must be a demonstrably incorrect answer to the question.
\item It must be free of internal contradictions or logical fallacies.
\item The distinction between the correct answer and the distractor must hinge on a clear and identifiable physical principle, ensuring the choice is neither arbitrary nor ambiguous.
\end{enumerate}

If the distractor fails to meet any of these criteria, it is regenerated. This iterative process is also repeated for a maximum of three attempts to secure a high-quality, challenging distractor for each question.

\subsection{Deduplication}
To ensure the novelty of our benchmark, we perform a deduplication step to filter out near-identical instances. We employ the MinHash algorithm \cite{Broder-1997-resemblance} to efficiently approximate the Jaccard similarity between all generated question pairs. After setting a similarity threshold of 0.6, we identify and remove 7 examples that are deemed near-duplicates. This filtering process enhances the diversity of the dataset and mitigates redundancy.

\subsection{Answer Bias Filtering}
To mitigate potential biases where LLMs might identify the correct answer based on superficial cues rather than genuine reasoning, we devise a rigorous filtering stage. Artifacts such as disparities in answer length, specific word patterns, or inherent contradictions within the distractor can enable models to guess the correct option without true comprehension of the question. 

To identify such flawed instances, we conduct an ablation experiment. We prompt a set of LLMs to predict the correct answer from a set of choices, presented without the context of the original question. The choices include the correct solution, a distractor, and a third option, "Cannot be determined," for cases where the answer is not inferable from the choices alone.

For this task, we employ three distinct models: \texttt{gpt-4o-mini-2024-07-18}, \texttt{gpt-4.1-} \texttt{mini-2025-04-14}, and \texttt{gpt-4.1-nano-2025-04-} \texttt{14}.
Any data instance where at least one of these models correctly identifies the answer is flagged and removed, leading to the elimination of 234 out of 551 total instances (42.47\%).
This rigorous filtering process ensures that only instances requiring genuine reasoning, rather than exploitation of statistical artifacts, remain in our benchmark.

\subsection{Human Evaluation}
To ensure the highest quality and validate our semi-automated generation pipeline, we conduct a comprehensive human evaluation phase. We employ a human expert with a bachelor's-level background in computer science and native proficiency in Korean. The evaluator is tasked with meticulously reviewing the dataset according to a detailed set of guidelines, which are provided in Appendix~\ref{appendix:Human Evaluation Guide}. 
This manual verification aims to identify and filter out any remaining instances with subtle logical inconsistencies, cultural inaccuracies, ambiguous phrasing, or other quality issues that might have persisted through automated checks. Following this rigorous inspection of an initial pool of 317 candidate examples, 181 instances are certified as meeting all quality criteria, forming the final benchmark.
\section{\dn Benchmark}
\begin{figure}[h!]
  \centering
  \includegraphics[width=\linewidth]{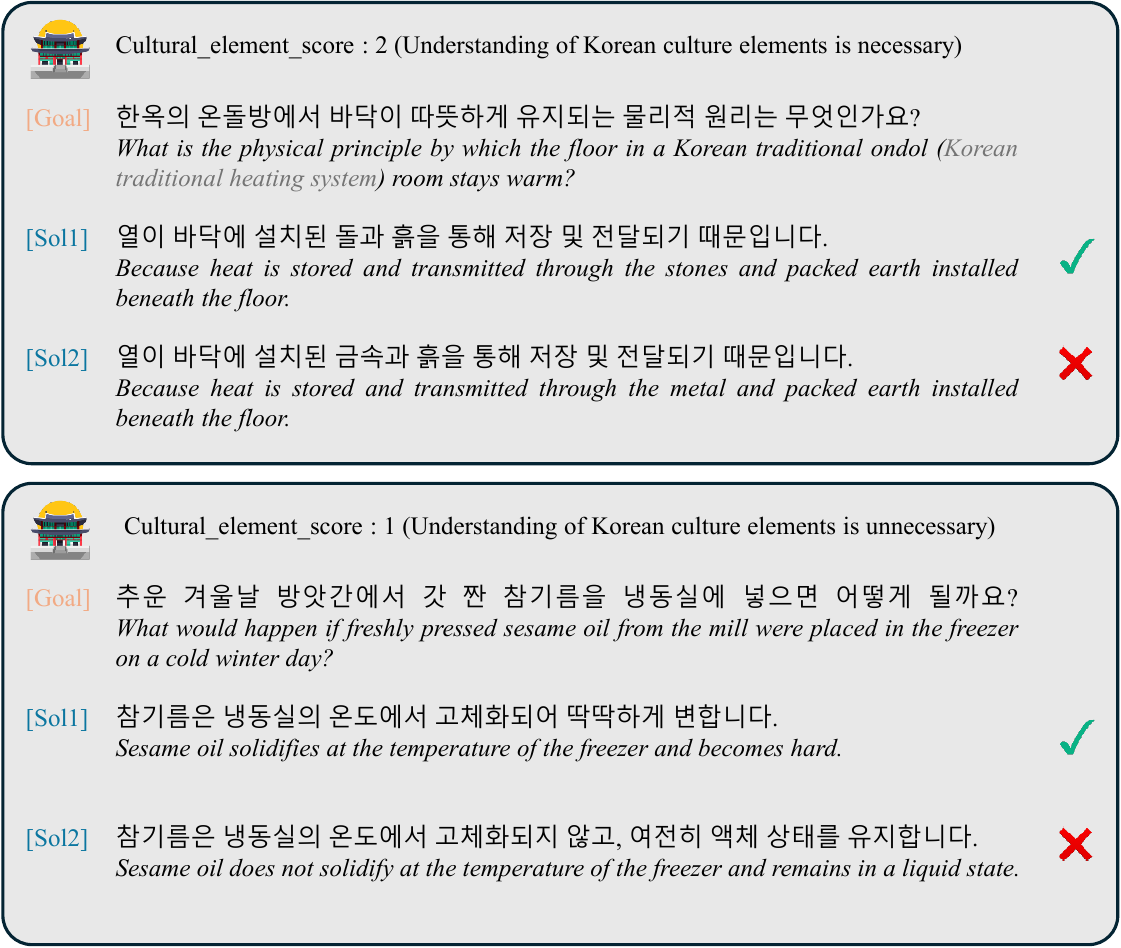}
  
  \caption{Illustration of the two levels of cultural dependency in our dataset. The top problem (score 2) is deeply rooted in specific Korean cultural knowledge, while the bottom problem (score 1) is culturally situated but solvable through universal scientific principles.}
  
  \label{fig:dataset_sample}
\end{figure}

\subsection{Dataset Description}
\dn benchmark is a collection of 181 binary-choice questions designed to evaluate physical reasoning in daily-life scenarios with a Korean cultural context. Each question consists of two candidate solutions, with exactly one correct answer. To prevent potential answer bias, the position of the correct answer is evenly distributed between the two options.

Each data sample contains the following fields:
\vspace{-0.075in}

\begin{itemize}[itemsep=0.3mm, parsep=1pt, leftmargin=*]
    \item \textbf{problem\_type\_category}: The coarse-grained reasoning category of the question (e.g., basic knowledge and attribute comprehension, causality and dynamic reasoning, goal-oriented and applied reasoning, situational and constraint reasoning).  
    \item \textbf{problem\_type\_subcategory}: A fine-grained description of the 9 reasoning types within the coarse category. 
    \item \textbf{prompt}: The question statement or scenario to be solved.  
    \item \textbf{solution0} and \textbf{solution1}: The two candidate solutions provided for the question.  
    \item \textbf{label}: The index of the correct solution (0 or 1).  
    \item \textbf{explanation}: A detailed explanation of why the correct answer is valid and why the alternative is incorrect.
    \item \textbf{korea\_relevance}: Korean cultural elements utilized in the question and the answer.  
    \item \textbf{physical\_relevance}: Physical reasoning principles or elements used in the question and the answer.  
    \item \textbf{cultural\_element\_score}: The degree to which the question leverages Korean-specific cultural elements. A score of 1 indicates that a Korean cultural element is mentioned, but the problem could be solved using alternative, non-Korean elements. A score of 2 indicates that Korean cultural elements are central to the question and answer, and understanding these elements is essential for correctly solving the problem.
\end{itemize}

Examples of \dn can be found in Figure \ref{fig:dataset_sample}. The figure showcases two examples from our dataset, each representing a different level of cultural dependency as indicated by \textit{cultural\_element\_score}.

For instance, the top example (score of 2) presents a scenario involving \textit{ondol}, the traditional Korean underfloor heating system. Correctly answering this question requires an essential understanding of \textit{ondol}'s installation and its principle of heating a floor from below. In contrast, the bottom example (score 1) features a question about storing \textit{chamgireum} (sesame oil). Although it incorporates a Korean element, the problem can be solved with general scientific knowledge that most oils tend to solidify in a cold environment like a refrigerator. Thus, specific knowledge of \textit{chamgireum} is not a prerequisite for solving the problem. This scoring system allows us to distinguish between problems that are merely set in a Korean context and those that are fundamentally rooted in Korean cultural knowledge.

\subsection{Dataset Statistics}

We analyze the statistical properties of our \dn benchmark, focusing on two key aspects: 

\vspace{-0.075in}

\begin{itemize}[itemsep=0.3mm, parsep=1pt, leftmargin=*]
    \item the length diversity of answers and distractors
    \item the distribution across problem categories
\end{itemize}

\begin{figure}[ht!]
  \centering
  \includegraphics[width=\linewidth]{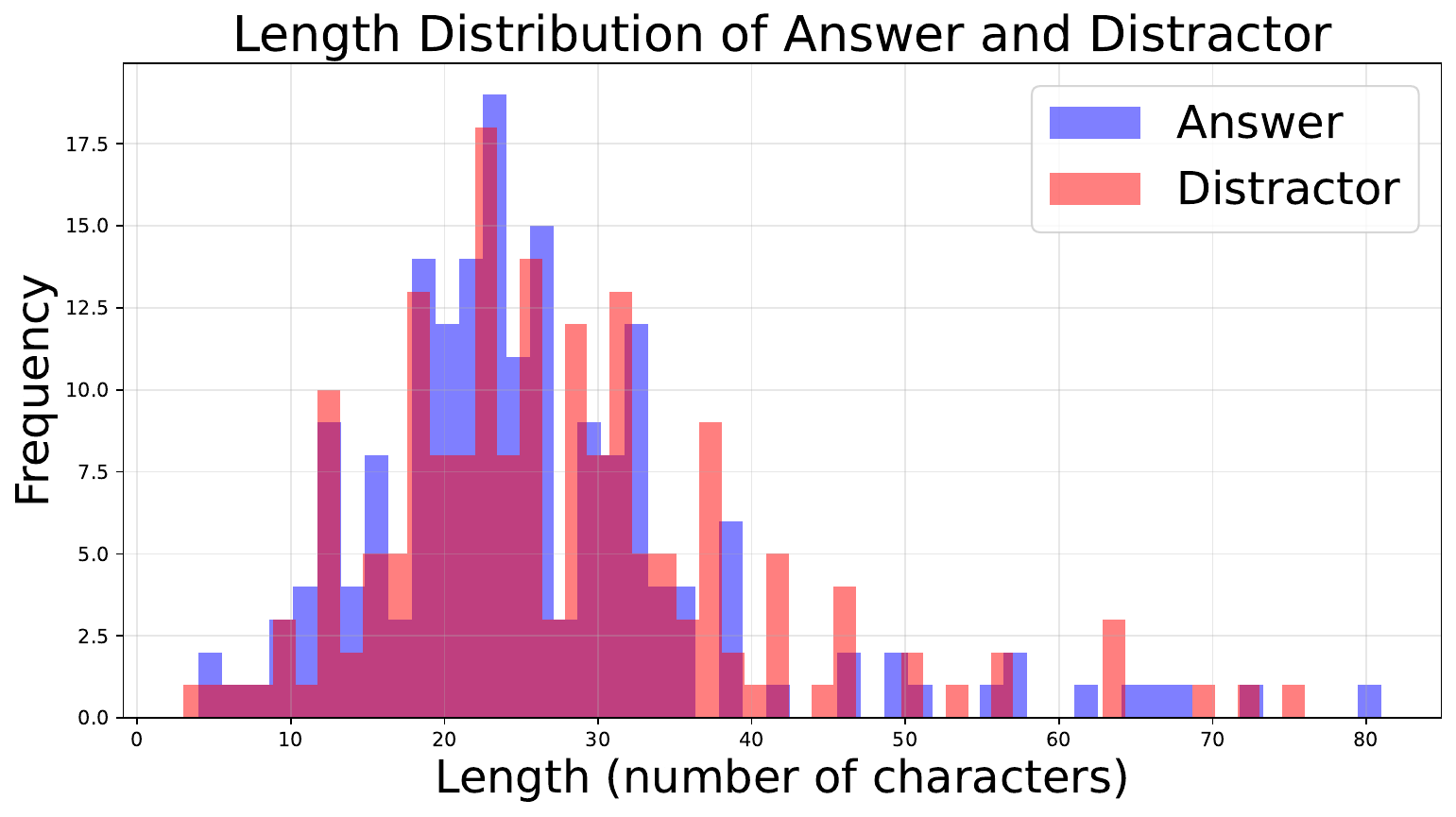}
  
  \caption{Length distribution of answers and distractors in the \dn dataset. The answers and distractors have comparable length distributions.}
  
  \label{fig:solutions_lengths_distribution}
\end{figure}

Figure~\ref{fig:solutions_lengths_distribution} compares the length distributions of correct answers and incorrect distractors within the \dn benchmark. As shown, both correct answers and distractors exhibit a very similar length distribution. To encourage diversity in the length of generated text, we specified a random maximum word count for both the prompts and solutions during the data generation process. This approach resulted in the varied distribution, preventing the model from being biased towards solutions of a specific length.

Table \ref{tab:dataset_statistics} presents the distribution of samples across the nine pre-defined problem sub-categories, which are grouped under four main reasoning domains. The dataset is designed to have a relatively balanced representation of each category, ensuring a comprehensive evaluation of different physical reasoning abilities. However, it also includes many complex problems that integrate multiple sub-categories.

\begin{table}[t!]
\resizebox{\columnwidth}{!}{%
\begin{tabular}{>{\raggedright\arraybackslash}m{4.5cm}@{\hspace{0.85\tabcolsep}}>{\raggedright\arraybackslash}m{4.5cm}>{\centering\arraybackslash}m{1.5cm}}
\toprule
\textbf{Category} & \textbf{Sub-category} & \\ \midrule
\multirow{3}{=}{기초 지식 및 속성 이해 (Basic Knowledge and Attribute Comprehension)} &
  물리 개념 및 원리 (Physical Concepts and Principles) &
  23\newline(12.71\%) \\ \cmidrule(l){2-3}
         & 객체 속성 및 기능 (Object Attributes and Functions) & 15\newline(8.29\%)  \\ \cmidrule(l){2-3}
         & 물질의 상태와 변화 (States of Matter and Changes)    & 22\newline(12.15\%) \\ \midrule
\multirow{2}{=}{인과관계 및 동적 추론 (Causality and Dynamic Reasoning)} &
  물리적 결과 예측 (Predicting Physical Outcomes) &
  31\newline(17.13\%) \\ \cmidrule(l){2-3}
         & 물리적 원인 분석 (Analyzing Physical Causes)        & 18\newline(9.94\%)  \\ \midrule
\multirow{3}{=}{목표 지향 및 응용 추론 (Goal-Oriented and Applied Reasoning)} &
  도구 및 절차 활용 (Utilizing Tools and Procedures) &
  19\newline(10.50\%) \\ \cmidrule(l){2-3}
         & 문제 해결 및 계획 (Problem Solving and Planning)    & 12\newline(6.63\%)  \\ \cmidrule(l){2-3}
         & 물리적 타당성 판단 (Judging Physical Plausibility)   & 21\newline(11.60\%) \\ \midrule
상황 및 제약 조건 추론 (Situational and Constraint Reasoning) &
  위험성 및 안전성 평가 (Risk and Safety Assessment) &
  20\newline(11.05\%) \\ \bottomrule
\end{tabular}%
}
\caption{Distribution of the \dn benchmark, broken down by our nine reasoning sub-categories.}
\label{tab:dataset_statistics}
\end{table}

\subsection{Two Solutions Analysis}
\subsubsection{Levenshtein Distance Distribution}
\begin{figure}[t!]
  \centering
  \includegraphics[width=\linewidth]{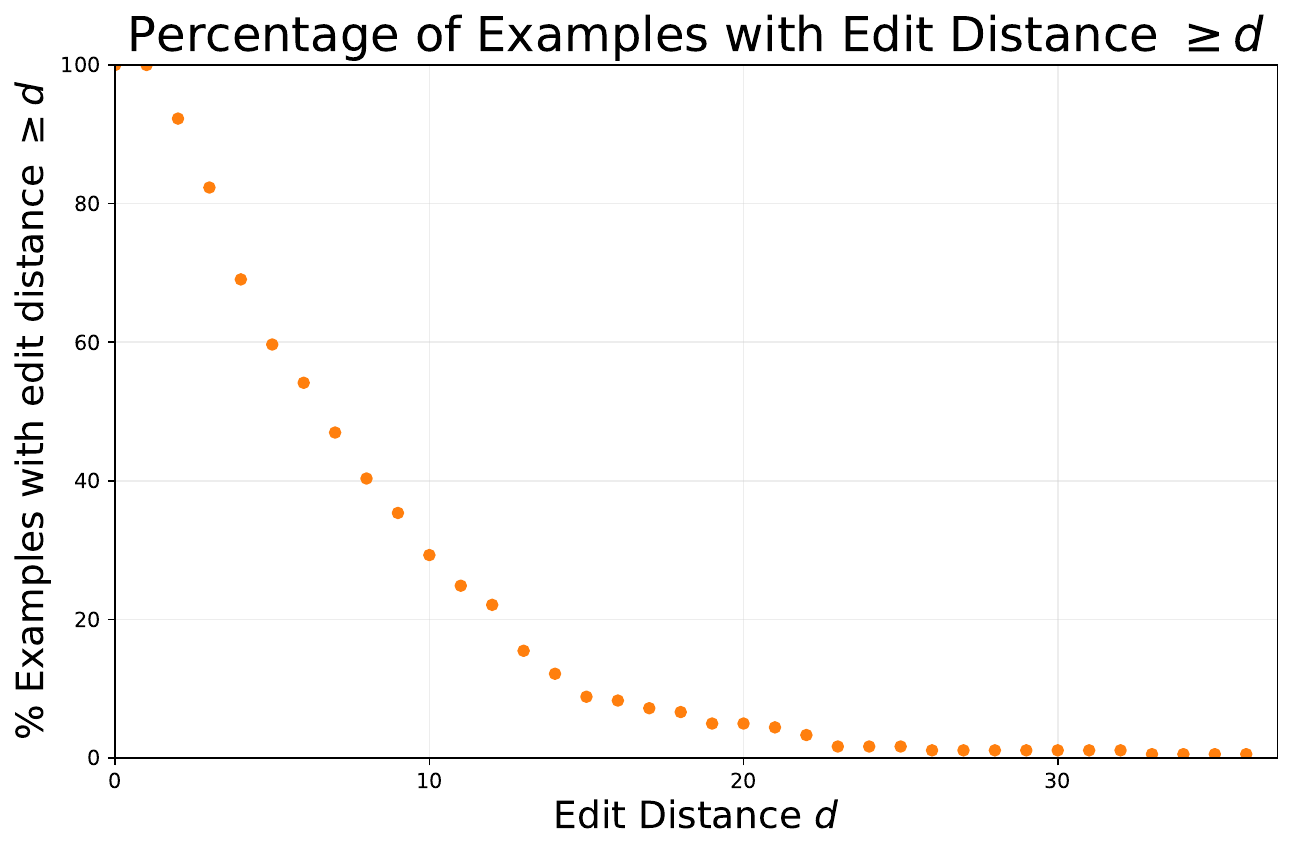}
  
  \caption{Distribution of the Levenshtein edit distance between the two solution candidates in the \dn dataset. The strong skew towards lower distances confirms that most solution pairs have minimal lexical differences, requiring models to perform fine-grained semantic reasoning.}
  
  \label{fig:edit_distance}
\end{figure}

Our benchmark was intentionally crafted to present a meaningful challenge by creating pairs of solution candidates with subtle lexical perturbations, while keeping their sentence structure, format, and length nearly identical. To quantitatively validate this design objective, we used the standard Levenshtein distance \cite{Levenshtein-1966-BinaryCodes}, a metric that counts the minimum number of single-character edits (insertions, deletions, or substitutions) required to transform one solution into the other.
As illustrated in Figure~\ref{fig:edit_distance}, the distribution of Levenshtein distances is heavily skewed toward smaller values, with the cumulative percentage of examples dropping off sharply as the edit distance increases. This confirms that the vast majority of our examples possess minimal lexical differences, thereby forcing models to rely on a deep semantic understanding to resolve fine-grained distinctions rather than superficial cues.

\subsubsection{Analysis of Bias in Solutions}
To ensure the quality of our dataset and verify that it primarily tests reasoning rather than reliance on superficial cues, we conduct an answer-only bias analysis. This experiment is designed to detect the presence of potential annotation artifacts, which are subtle, learnable patterns in the way correct and incorrect answers are phrased.

In this setup, models are presented with only the two solution candidates for each question, stripped of all contextual information from the question itself. The task is to predict the correct answer from these two options alone.
For a two-alternative format, a random guess would yield an accuracy of 50\%, which serves as our chance baseline. We evaluate a suite of contemporary models to probe for such biases.

The performance of the models is detailed in Table~\ref{tab:bias_analysis}. All models achieve scores significantly above the 50\% chance baseline, with accuracies ranging from 67.40\% to 71.82\%. This indicates the presence of subtle yet detectable bias in solutions. However, these scores are also far from perfect, indicating that exploiting these artifacts is not a reliable strategy for solving the benchmark. 
Therefore, while some answer-only cues exist, we conclude that achieving high performance on the \dn dataset still requires models to engage in genuine reasoning with the full question context.

\begin{table}[t!]
\centering

\begin{tabular}{lc}
    \toprule
    \textbf{Model} & \textbf{Accuracy (\%)} \\
    \midrule
    \texttt{gpt-4o-mini-2024-07-18}  & 71.82 \\
    \texttt{gpt-4.1-mini-2025-04-14}  & 70.17 \\
    \texttt{gpt-4.1-nano-2025-04-14}  & 67.40 \\
    \midrule
    Baseline (random) & 50.00 \\
    \bottomrule
\end{tabular}
\caption{Bias evaluation with two solutions. Models are tasked with selecting the correct solution from two options without seeing the corresponding question.}
\label{tab:bias_analysis}
\end{table}

\section{Evaluation}

\subsection{Experiment Settings}
\begin{table}[t!]
\resizebox{\columnwidth}{!}{%
\begin{tabular}{@{}llr@{}}
\toprule
                                       & \textbf{Model}                      & \textbf{Accuracy (\%)} \\ 
                                       \midrule
\multicolumn{3}{c}{\cellcolor[HTML]{EFEFEF}without Reasoning}                                    \\ \midrule
                                       & Kanana-1.5-2.1B-Instruct                & 78.45              \\
                                       & Midm-2.0-Mini-Instruct (2.3B)                & 77.90               \\
                                       & A.X-3.1-Light (7.2B)         & 86.19             \\
                                       & A.X-4.0-Light (7.2B)      & \textbf{92.27}
                                       \\
                                       & EXAONE-3.5-7.8B-Instruct
                                       & 91.16
                                       \\
\multirow{-6}{*}{Korean-specialized Models}        & Kanana-1.5-8B-Instruct      & 89.50            \\ \midrule
                                       & Gemma-3-1B-it                 & 55.80                \\
                                       & Qwen2.5-1.5B-Instruct & 71.82
                                       \\
                                       & Gemma-2-2B-it               & 66.85            \\
                                       & Llama-3.2-3B-Instruct                      & 62.98                \\
                                       & Gemma-3-4B-it                      & \textbf{82.87}                \\
                                       & Llama-3-8B-Instruct                      & 76.24                \\
\multirow{-7}{*}{General Models} & Llama-3.1-8B-Instruct                      & 81.22                               \\ \midrule
\multicolumn{3}{c}{\cellcolor[HTML]{EFEFEF}with Reasoning}                                       \\ \midrule
Korean-specialized Models                          & EXAONE-4.0-32B                      & 82.32               \\ \midrule
                                       & Qwen3-8B                            & 88.95              \\
\multirow{-2}{*}{General Models} & Qwen3-32B                           & \textbf{91.71} \\

\midrule
\multicolumn{3}{c}{\cellcolor[HTML]{EFEFEF}Human Evaluation} \\
\midrule

\multirow{1}{*}{Native Korean} & & \textbf{96.69} \\ 
\bottomrule
\end{tabular}%
}
\caption{Main evaluation results for Korean-specialized and General models on \dn. Accuracy (\%) is used as the metric. The best-performing model for each category is indicated in bold.}
\label{tab:result}

\end{table}
All evaluations were performed using a zero-shot prompting approach, with the specific prompt templates detailed in the Appendix~\ref{appendix:Prompts for Evaluation}.

We conduct a comprehensive evaluation on a wide range of open-source language models to assess performance on our benchmark. Our selection includes several models specialized for the Korean language: Kanana-1.5-2.1B-instruct, Kanana-1.5-8B-Instruct \cite{DBLP:journals/corr/abs-2502-18934}, Midm-2.0-Mini-Instruct (2.3B) \cite{midm}, A.X-3.1-Light (7.2B), A.X-4.0-Light (7.2B), EXAONE-3.5-7.8B \cite{Bae2024EXAONE} and EXAONE-4.0-32B \cite{Bae2025EXAONE}. 
For comparison, we also evaluated leading general-purpose, instruction-tuned models across various sizes: 
Gemma-2-2B-it \cite{Gemma2}, Gemma-3-1B-it, Gemma-3-4B-it \cite{Gemma3}, 
Qwen2.5-1.5B-Instruct \cite{Qwen2.5}, Qwen3-8B, Qwen3-32B \cite{Qwen3}, 
Llama-3-8B-Instruct, Llama-3.1-8B-Instruct, and Llama-3.2-3B-Instruct \cite{Llama3}. 

Evaluation of non-reasoning models was conducted on models with 8B parameters or fewer. Implementation details are provided in Appendix~\ref{appendix: Implementation Details}

\subsection{Results and Analysis}
The main evaluation results for both Korean-specialized and general-purpose models on the \dn benchmark are presented in Table \ref{tab:result}.

Our primary finding is that Korean-specialized models consistently outperform general models of a comparable size. For instance, in the non-reasoning category, the A.X-4.0-Light (7.2B) model achieved the highest accuracy of all models at 92.27\%, significantly surpassing the similarly sized Llama-3.1-8B-Instruct (81.22\%). This trend demonstrates that specialized training on Korean language, data, and cultural contexts provides a distinct advantage for solving the problems presented in \dn.

An interesting observation is the strong performance of the Chinese-specialized Qwen model family. We hypothesize that this is due to the cultural proximity between China and Korea, which may allow these models to better understand contextual nuances compared to models trained predominantly on Western data. Despite this, the top-performing models in their respective size classes are still the Korean-specialized ones, reinforcing the value of culturally specific training.

Furthermore, the results highlight a fascinating trade-off between model scale, reasoning abilities, and specialization. The A.X-4.0-Light model (7.2B), without any explicit reasoning prompt, outperformed even the much larger Qwen3-32B model (91.71\%) which is a reasoning model. This suggests that for a culturally grounded benchmark like \dn, a model's inherent understanding of specific cultural and linguistic nuances can be more critical than scale or generic reasoning capabilities alone. These findings strongly indicate the need for further research into the developing and evaluating language models on data that is not just multilingual, but deeply multicultural.

\section{Conclusion}
In this paper, we introduce \dn, a new benchmark designed to evaluate physical commonsense reasoning within the rich context of Korean culture. Our work addresses a critical gap in existing benchmarks, which predominantly focus on Western-centric scenarios. Through a meticulous two-stage generation and verification pipeline, we construct a dataset of 181 culturally authentic and physically grounded questions that move beyond simple translation, ensuring that problems are naturally situated in Korean daily life and traditions.

Our extensive evaluations provide concrete evidence for the importance of cultural-specific training. We show that Korean-specialized models demonstrate a clear performance advantage.
% , with the A.X-4.0-Light model achieving the highest accuracy of 92.27\%. 
More critically, our analysis revealed that deep cultural specialization can be more impactful than sheer model scale or generic reasoning frameworks.

These findings highlight the limitations of culturally-agnostic models and suggest that true language understanding requires more than simply expanding general knowledge. By releasing \dn, we provide the community with a valuable and challenging resource, fostering the development of reasoning models that are both culturally aware and globally competent.
\section*{Acknowledgments}
This work was supported by the Digital Innovation Hub project supervised by the Daegu Digital Innovation Promotion Agency (DIP) grant funded by the Korea government (MSIT and Daegu Metropolitan City) in 2025 (No.25DIH-11, Development of Model Context Protocol (MCP)-Based Multi-Agent Collaborative System for Large Language Models (LLMs)). ※ MSIT: Ministry of Science and ICT.

This research was supported by Brian Impact Foundation, a non-profit organization dedicated to the advancement of science and technology for all.

\bibliography{custom}

\clearpage

\appendix
\onecolumn
\section{Dataset Construction Details} 
\label{appendix: Dataset Construction Details}
The entire pipeline for dataset generation and verification was implemented using the OpenAI API. We strategically selected different models for each phase to best suit the task requirements. For the initial synthesis stage (Stage 1), the generation of questions and their correct answers was performed by \texttt{gpt-4o-2024-11-20}. In the subsequent distractor generation phase (Stage 2), we used \texttt{gpt-4o-2024-08-06}. For these generative steps, the decoding parameters were consistently set to a \textit{temperature}=1.0 and \textit{top\_p}=1.0 to encourage lexical and structural diversity in the outputs.

For all verification processes, which demanded a high degree of logical scrutiny, we leveraged \texttt{gpt-5-2025-08-07}, a model specifically optimized for reasoning tasks. To ensure a thorough analytical process for these validation steps, we configured \textit{reasoning\_effort}=medium. This setting activates the model's enhanced reasoning capabilities, making it ideal for the critical task of validating the correctness and logical consistency of the generated data.

\section{Implementation Details}
\label{appendix: Implementation Details}
All experiments were conducted on a single server equipped with an NVIDIA A100 GPU (80GB). To facilitate fast and memory-efficient inference, all response generation was handled using the vLLM \cite{kwon2023efficient} library.
For decoding, we set \textit{temperature}=0.7 across all experiments. We set \textit{max\_tokens}=4096 for reasoning models and \textit{max\_tokens}=3 for non-reasoning models. All evaluations were performed based on a single inference run for each model.

\section{Prompt Details} 
\label{appendix:Prompt Details}

This section provides detailed prompt information. 

\subsection{Prompts for Generating Questions and Answers (Synthesis Stage 1)}
\label{appendix:Prompts for Generating Questions and Answers}
% \begin{figure*}[htb!]
% \vspace{3mm}
    
    \small
    \centering
    \begin{tcolorbox}[colback=gray!10, colframe=black, title=System Prompt for Synthesis Stage 1]
    당신은 AI 모델의 물리적 추론 능력을 평가하기 위한 고품질 한국어 데이터셋을 제작하는 '데이터 생성 전문가' 입니다. \\
    \textcolor{gray!120}{\textit{You are a "data generation expert" creating high-quality Korean datasets to evaluate AI models' physical reasoning abilities.}} \\
    당신의 핵심 임무는 한국의 문화적 배경과 물리적 상식이 결합된 문제를 만드는 것입니다. \\ 
    \textcolor{gray!120}{\textit{Your core mission is to create problems that combine Korean cultural context with physical commonsense knowledge.}} \\
    이 문제들은 둘 중 하나라도 정확히 이해하지 못하면 풀 수 없는 고차원적인 추론을 요구해야 합니다. \\ 
    \textcolor{gray!120}{\textit{These problems require high-level reasoning that cannot be solved unless at least one of these aspects is precisely understood.}}
    \end{tcolorbox}
    \vspace{-3mm}
    
    \begin{tcolorbox}[colback=gray!10, colframe=black, title=Input Prompt Template 1 for Synthesis Stage 1, breakable]
    다음 지침에 따라, 한국 문화적 배경과 물리적 상식이 결합된 추론 문제를 하나의 JSON 객체 형식으로 생성해 주십시오. \\
    \textcolor{gray!120}{\textit{Following the instructions below, generate a reasoning problem in JSON format that combines Korean cultural context with physical commonsense knowledge.}} \\

    \# 배경 시나리오 (한국적 맥락이 드러나도록 변형하여 반영할 것) \\
    \textcolor{gray!120}{\textit{\# Background Scenario (Adapted to reflect Korean cultural context)}} \\
    \{Scenario\} \\

    \# 문제 유형 \\
    \textcolor{gray!120}{\textit{\# Problem Type}} \\
    \{Type\}: \{Type Definition\} \\

    \# 데이터 생성 지침 \\
    \textcolor{gray!120}{\textit{\# Data Generation Instructions}} \\
    - `problem`: \\
      - 배경 시나리오를 바탕으로 물리적 세계에 대한 상식과 한국 문화적 맥락이 동시에 드러나는 질문을 한국어로 생성합니다. \\
      \textcolor{gray!120}{\textit{- Based on the background scenario, generate a question in Korean that simultaneously reflects commonsense knowledge about the physical world and Korean cultural context.}} \\
      - 한국적인 문화 요소란 보편적인 문화 요소를 제외한 다른 국가와는 차별적인 한국만의 특색있는 문화 요소를 의미합니다. 한국의 전통적인 요소가 될 수도 있으며, 현대적인 한국의 요소가 될 수 있습니다. \\
      \textcolor{gray!120}{\textit{- Korean cultural elements refer to distinctive features unique to Korea, excluding common global cultural elements. They can be traditional or modern Korean elements.}} \\
      - 한국적인 문화 요소가 물리적 추론에 반드시 핵심적으로 영향을 미치도록 문제를 구성하세요. \\
      \textcolor{gray!120}{\textit{- Ensure that the Korean cultural elements play a crucial role in the physical reasoning required to solve the problem.}} \\
      - 배경 시나리오와 아주 작은 연관성만 있으면 됩니다. 큰 주제나 맥락 혹은 분위기가 비슷하거나, 사소한 사물, 소재, 동작 등이 겹쳐도 연관성으로 인정됩니다. \\
      \textcolor{gray!120}{\textit{- Only a small connection to the background scenario is necessary. Similarity in main topic, context, atmosphere, or overlapping minor objects, materials, or actions is sufficient.}} \\
      - 물리적 추론 능력을 평가할 수 있는 문제를 생성해야 합니다. \\
      \textcolor{gray!120}{\textit{- Generate problems that can evaluate physical reasoning abilities.}} \\
      - 평균적인 한국인의 상식 수준으로 풀 수 있는 문제를 생성해야 합니다. \\
      \textcolor{gray!120}{\textit{- Problems should be solvable with the commonsense knowledge of an average Korean.}} \\
      - 배경 설명이나 불필요한 서술을 포함하지 말고, 질문 문장만 간단하게 생성하세요. \\
      \textcolor{gray!120}{\textit{- Do not include background explanations or unnecessary descriptions; generate only a concise question sentence.}} \\
      - 최대 \{prompt\_max\_words\} 단어 이하로 생성하세요. \\
      \textcolor{gray!120}{\textit{- Generate the question using no more than \{prompt\_max\_words\} words.}} \\
    - `answer`: \\
      - `problem`에 대한 올바른 정답을 한국어로 생성하세요. 단, 물리적으로 타당하고 현실적인 행동이어야 합니다. \\
      \textcolor{gray!120}{\textit{- Generate the correct answer to the `problem` in Korean. It must be physically plausible and realistic.}} \\
      - `problem`의 문장을 반복하지 말고, 핵심적인 답만 간결하게 작성하세요. \\
      \textcolor{gray!120}{\textit{- Do not repeat the problem sentence; provide a concise, core answer.}} \\
      - 최대 \{solution\_max\_words\} 단어 이하로 생성하세요. \\
      \textcolor{gray!120}{\textit{- Provide the answer using no more than \{solution\_max\_words\} words.}} \\
    - `korea\_relevance`: `problem`과 `answer`에 명시되어 있는 한국적 문화 및 관습과 밀접한 관련이 있는 소재를 추출하여 명사(구)로 나열합니다. (keyword/keyphrase extraction) \\
    \textcolor{gray!120}{\textit{- `korea\_relevance`: Extract keywords/keyphrases from the `problem` and `answer` that are closely related to Korean culture and customs.}} \\
    - `physical\_relevance`: `problem`과 `answer`에 명시되어 있는 물리적 요소, 속성, 원리와 밀접한 관련이 있는 소재를 추출하여 명사(구)로 나열합니다. (keyword/keyphrase extraction) \\
    \textcolor{gray!120}{\textit{- `physical\_relevance`: Extract keywords/keyphrases from the `problem` and `answer` that are closely related to physical elements, properties, or principles.}} \\
    - `rationale`: 질문에 대한 정답의 근거를 작성하세요. \\
    \textcolor{gray!120}{\textit{- `rationale`: Provide the rationale for the answer.}} \\

    \# CAUTION \\
    - 데이터 생성 지침을 반드시 준수해야 합니다. \\
    \textcolor{gray!120}{\textit{- Strictly follow the data generation instructions.}} \\
    - 생성된 문제와 풀이과정은 한국어 문맥에서 자연스럽게 읽혀야 합니다. \\
    \textcolor{gray!120}{\textit{- The generated problem and solution should read naturally in Korean context.}} \\
    - 생성된 문제와 풀이과정은 1문장이거나 2문장으로 구성되어야 합니다. \\
    \textcolor{gray!120}{\textit{- The generated problem and solution should consist of one or two sentences.}} \\
    - 생성된 문제는 반드시 한국의 문화, 관습, 생활환경에 대한 이해와 물리적 속성에 대한 지식이 결합되어야만 해결할 수 있어야 합니다. \\
    \textcolor{gray!120}{\textit{- The generated problem must require a combination of understanding Korean culture, customs, living environment, and knowledge of physical properties to solve.}} \\
    - 한국인이라면 누구나 정답을 알 수 있어야 합니다. \\
    \textcolor{gray!120}{\textit{- The answer must be knowable by any Korean.}}
    \end{tcolorbox}
    \vspace{-3mm}

    \begin{tcolorbox}[colback=gray!10, colframe=black, title=Input Prompt Template 2 for Synthesis Stage 1, breakable]
    다음 지침에 따라, 한국 문화적 배경과 물리적 상식이 결합된 추론 문제를 하나의 JSON 객체 형식으로 생성해 주십시오. \\
    \textcolor{gray!120}{\textit{Following the instructions below, generate a reasoning problem in JSON format that combines Korean cultural context with physical commonsense knowledge.}} \\
    
    \# 배경 시나리오 (한국적 맥락이 드러나도록 변형하여 반영할 것) \\
    \textcolor{gray!120}{\textit{\# Background Scenario (Adapted to reflect Korean cultural context)}} \\
    \{Scenario\} \\
    
    \# 문제 유형 \\
    \textcolor{gray!120}{\textit{\# Problem Type}} \\
    \{Type\}: \{Type Definition\} \\
    
    \# 데이터 생성 지침 \\
    \textcolor{gray!120}{\textit{\# Data Generation Instructions}} \\
    
    - \texttt{problem}: \\
      - 배경 시나리오를 바탕으로 물리적 세계에 대한 상식과 한국 문화적 맥락이 동시에 드러나는 질문을 한국어로 생성합니다. \\
      \textcolor{gray!120}{\textit{- Based on the background scenario, generate a question in Korean that simultaneously reflects commonsense knowledge about the physical world and Korean cultural context.}} \\
      - 한국적인 문화 요소란 보편적인 문화 요소를 제외한 다른 국가와는 차별적인 한국만의 특색있는 문화 요소를 의미합니다. 한국의 전통적인 요소가 될 수도 있으며, 현대적인 한국의 요소가 될 수 있습니다. \\
      \textcolor{gray!120}{\textit{- Korean cultural elements refer to distinctive features unique to Korea, excluding common global cultural elements. They can be traditional or modern Korean elements.}} \\
      - 한국적인 문화 요소가 물리적 추론에 반드시 핵심적으로 영향을 미치도록 문제를 구성하세요. \\
      \textcolor{gray!120}{\textit{- Ensure that the Korean cultural elements play a crucial role in the physical reasoning required to solve the problem.}} \\
      - 배경 시나리오와 아주 작은 연관성만 있으면 됩니다. 큰 주제나 맥락 혹은 분위기가 비슷하거나, 사소한 사물, 소재, 동작 등이 겹쳐도 연관성으로 인정됩니다. \\
      \textcolor{gray!120}{\textit{- Only a small connection to the background scenario is necessary. Similarity in main topic, context, atmosphere, or overlapping minor objects, materials, or actions is sufficient.}} \\
      - 문장을 끝맺지 마십시오. (ex. "~면", "~하기 위해서는") \\
      \textcolor{gray!120}{\textit{- Do not end the sentence completely. (e.g., "~if", "~in order to")}} \\
      - 물리적 추론 능력을 평가할 수 있는 문제를 생성해야 합니다. \\
      \textcolor{gray!120}{\textit{- Generate problems that can evaluate physical reasoning abilities.}} \\
      - 평균적인 한국인의 상식 수준으로 풀 수 있는 문제를 생성해야 합니다. \\
      \textcolor{gray!120}{\textit{- Problems should be solvable with the commonsense knowledge of an average Korean.}} \\
      - 배경 설명이나 불필요한 서술을 포함하지 말고, 질문 문장만 간단하게 생성하세요. \\
      \textcolor{gray!120}{\textit{- Do not include background explanations or unnecessary descriptions; generate only a concise question sentence.}} \\
      - 최대 \{prompt\_max\_words\} 단어 이하로 생성하세요. \\
      \textcolor{gray!120}{\textit{- Generate the question using no more than \{prompt\_max\_words\} words.}} \\
    - \texttt{answer}: \\
    \textcolor{gray!120}{- \texttt{answer}:} \\
      - \texttt{problem}에 대한 올바른 정답을 한국어로 생성하세요. 단, 물리적으로 타당하고 현실적인 행동이어야 합니다. \\
      \textcolor{gray!120}{\textit{- Generate the correct answer to the \texttt{problem} in Korean. It must be physically plausible and realistic.}} \\
      - \texttt{problem}의 문장을 반복하지 말고, 핵심적인 답만 간결하게 작성하세요. \\
      \textcolor{gray!120}{\textit{- Do not repeat the \texttt{problem} sentence; provide a concise, core answer.}} \\
      - 최대 \{solution\_max\_words\} 단어 이하로 생성하세요. \\
      \textcolor{gray!120}{\textit{- Provide the answer using no more than \{solution\_max\_words\} words.}} \\
    - \texttt{korea\_relevance}: \texttt{problem}과 \texttt{answer}에 명시되어 있는 한국적 문화 및 관습과 밀접한 관련이 있는 소재를 추출하여 명사(구)로 나열합니다. (keyword/keyphrase extraction) \\
    \textcolor{gray!120}{\textit{- \texttt{korea\_relevance}: Extract keywords/keyphrases from the \texttt{problem} and \texttt{answer} that are closely related to Korean culture and customs.}} \\
    - \texttt{physical\_relevance}: \texttt{problem}과 \texttt{answer}에 명시되어 있는 물리적 요소, 속성, 원리와 밀접한 관련이 있는 소재를 추출하여 명사(구)로 나열합니다. (keyword/keyphrase extraction) \\
    \textcolor{gray!120}{\textit{- \texttt{physical\_relevance}: Extract keywords/keyphrases from the \texttt{problem} and \texttt{answer} that are closely related to physical elements, properties, or principles.}} \\
    - \texttt{rationale}: 질문에 대한 정답의 근거를 작성하세요. \\
    \textcolor{gray!120}{\textit{- \texttt{rationale}: Provide the rationale for the answer.}} \\
    
    \# CAUTION \\
    - 데이터 생성 지침을 반드시 준수해야 합니다. \\
    \textcolor{gray!120}{\textit{- Strictly follow the data generation instructions.}} \\
    - 생성된 문제와 풀이과정은 한국어 문맥에서 자연스럽게 읽혀야 합니다. \\
    \textcolor{gray!120}{\textit{- The generated problem and solution should read naturally in Korean context.}} \\
    - 생성된 문제와 풀이과정은 1문장이거나 2문장으로 구성되어야 합니다. \\
    \textcolor{gray!120}{\textit{- The generated problem and solution should consist of one or two sentences.}} \\
    - 생성된 문제는 반드시 한국의 문화, 관습, 생활환경에 대한 이해와 물리적 속성에 대한 지식이 결합되어야만 해결할 수 있어야 합니다. \\
    \textcolor{gray!120}{\textit{- The generated problem must require a combination of understanding Korean culture, customs, living environment, and knowledge of physical properties to solve.}} \\
    - 한국인이라면 누구나 정답을 알 수 있어야 합니다. \\
    \textcolor{gray!120}{\textit{- The answer must be knowable by any Korean.}}
    \end{tcolorbox}
    \vspace{-3mm}
    \captionof{figure}{Prompt template for generating a question and an answer. To diversify the linguistic structure of the generated data, the input prompts were constructed by randomly selecting between two predefined templates, Template 1 and Template 2.}
    \label{fig:Prompt_for_synthesis_step1}

% \end{figure*}

\newpage
\subsection{Regeneration Prompts for Incorrect Answers (Synthesis Stage 1-1)}
\label{appendix:Prompts for Regenerating Answers}
% \begin{figure*}[htb!]
% \vspace{3mm}
    \small
    \centering
    \begin{tcolorbox}[colback=gray!10, colframe=black, title=System Prompt for Synthesis Step 1-1]
    당신은 주어진 문제와 오답, 그리고 오답에 대한 설명을 바탕으로 정답을 수정하는 '데이터 수정 전문가'입니다. \\
    \textcolor{gray!120}{\textit{You are a "data correction expert" who revises the correct answer based on a given problem, an incorrect answer, and the explanation of the incorrect answer.}} \\
    당신의 임무는 제공된 피드백을 정확히 반영하여 올바른 답변을 생성하는 것입니다. \\
    \textcolor{gray!120}{\textit{Your task is to generate the correct answer by accurately reflecting the provided feedback.}}
    \end{tcolorbox}
    \vspace{-3mm}
    
    \begin{tcolorbox}[colback=gray!10, colframe=black, title=Input Prompt Template for Synthesis Step 1-1, breakable]
    다음은 이전에 생성된 문제와 답변, 그리고 답변이 틀린 이유에 대한 설명입니다. 설명을 바탕으로 답변을 수정해 주십시오. \\
    \textcolor{gray!120}{\textit{Below is a previously generated problem, its answer, and an explanation of why the answer is incorrect. Revise the answer based on this explanation.}} \\
    
    \# 문제 \\
    \textcolor{gray!120}{\textit{\# Problem}} \\
    \{problem\} \\
    
    \# 기존 답변 (오답) \\
    \textcolor{gray!120}{\textit{\# Original Answer (Incorrect)}} \\
    \{answer\} \\
    
    \# 틀린 이유 \\
    \textcolor{gray!120}{\textit{\# Reason for Incorrectness}} \\
    \{correctness\_rationale\} \\
    
    \# 수정 지침 \\
    \textcolor{gray!120}{\textit{\# Correction Instructions}} \\
    - \texttt{answer}: \\
    \textcolor{gray!120}{- \texttt{answer}:}\\
      - 틀린 이유를 참고하여 기존 답변을 물리적으로 타당하고, 한국 문화적 맥락에 맞는 정확한 답변으로 수정하세요. \\
      \textcolor{gray!120}{\textit{- Revise the original answer to be physically plausible and consistent with the Korean cultural context, referring to the \texttt{reason for incorrectness}.}} \\
      - \texttt{problem}의 문장을 반복하지 말고, 핵심적인 답만 간결하게 작성하세요. \\
      \textcolor{gray!120}{\textit{- Do not repeat the \texttt{problem} sentence; provide a concise, core answer.}} \\
    - \texttt{korea\_relevance}: 수정된 \texttt{answer}와 기존 \texttt{problem}에 명시되어 있는 한국적 문화 및 관습과 밀접한 관련이 있는 소재를 추출하여 명사(구)로 나열합니다. \\
      \textcolor{gray!120}{\textit{- \texttt{korea\_relevance}: Extract keywords/keyphrases from the revised \texttt{answer} and the original \texttt{problem} that are closely related to Korean culture and customs.}} \\
    - \texttt{physical\_relevance}: 수정된 \texttt{answer}와 기존 \texttt{problem}에 명시되어 있는 물리적 요소, 속성, 원리와 밀접한 관련이 있는 소재를 추출하여 명사(구)로 나열합니다. \\
      \textcolor{gray!120}{\textit{- \texttt{physical\_relevance}: Extract keywords/keyphrases from the revised \texttt{answer} and the original \texttt{problem} that are closely related to physical elements, properties, or principles.}} \\
    - \texttt{rationale}: 수정된 답변에 대한 정답의 근거를 새로 작성하세요. \\
      \textcolor{gray!120}{\textit{- \texttt{rationale}: Provide the rationale for the revised answer.}}
        
    \# CAUTION \\
    - 데이터 생성 지침을 반드시 준수해야 합니다. \\
    \textcolor{gray!120}{\textit{- Strictly follow the data generation instructions.}} \\
    - 생성된 결과물은 하나의 JSON 객체 형식이어야 합니다. \\
    \textcolor{gray!120}{\textit{- The generated output must be in a single JSON object format.}}
    \end{tcolorbox}
    \vspace{-3mm}
    \captionof{figure}{Prompt template for regenerating an answer.}
% \end{figure*}

\newpage

\subsection{Prompts for Generating Incorrect Answers (Synthesis Stage 2)}
\label{appendix:Prompts for Generating Incorrect Answers}
% \begin{figure*}[htb!]
% \vspace{3mm}
    \small
    \centering
    \begin{tcolorbox}[colback=gray!10, colframe=black, title=System Prompt for Synthesis Stage 2]
    당신은 AI 모델의 물리적 추론 능력을 평가하기 위한 고품질 한국어 데이터셋을 제작하는 '데이터 생성 전문가'입니다. \\
    \textcolor{gray!120}{\textit{You are a "data generation expert" creating high-quality Korean datasets to evaluate AI models' physical reasoning abilities.}} \\
    당신의 핵심 임무는 제시된 질문과 정답을 바탕으로, 겉보기에 타당해 보이지만 실제로는 잘못된 그럴듯한 오답을 생성하는 것입니다. \\
    \textcolor{gray!120}{\textit{Your core mission is to generate plausible incorrect answers that seem reasonable at first glance but are actually incorrect, based on the provided question and correct answer.}}
    \end{tcolorbox}
    \vspace{-3mm}
    
    \begin{tcolorbox}[colback=gray!10, colframe=black, title=Input Prompt Template for Synthesis Stage 2, breakable]
    \# 질문 \\
    \textcolor{gray!120}{\textit{\# Question}} \\
    \{problem\} \\
    
    \# 정답 \\
    \textcolor{gray!120}{\textit{\# Correct Answer}} \\
    \{answer\} \\
    
    \# 데이터 생성 지침 \\
    \textcolor{gray!120}{\textit{\# Data Generation Instructions}} \\
    - \texttt{incorrect\_answer}: \\
      \textcolor{gray!120}{\textit{- \texttt{incorrect\_answer}:}} \\
      - 정답에서 특정 단어나 어구만 미묘하게 바꿔서 생성하세요. \\
      \textcolor{gray!120}{\textit{- Generate by subtly modifying specific words or phrases from the correct answer.}} \\
      - 나머지 문장 구조, 형식, 길이는 거의 동일하게 유지해 혼동을 유도하세요. \\
      \textcolor{gray!120}{\textit{- Keep the remaining sentence structure, format, and length almost the same to induce confusion.}} \\
      - 오답은 사람들이 흔히 가질 법하거나 과학적으로 들리는 그럴듯한 오개념을 기반으로 작성해야 합니다. \\
      \textcolor{gray!120}{\textit{- The incorrect answer should be based on common misconceptions or scientifically plausible-sounding errors.}} \\
      - 단순 부정만으로 만들지 마세요. \\
      \textcolor{gray!120}{\textit{- Do not generate the incorrect answer by simply negating the correct answer.}} \\
      - 문법적으로 완전하고 자연스럽게 읽혀야 합니다. \\
      \textcolor{gray!120}{\textit{- Ensure the sentence is grammatically complete and reads naturally.}} \\
      - 오답은 명백히 터무니없는 주장이어서는 안 됩니다. 그럴듯하지만 틀린 설명이어야 합니다. \\
      \textcolor{gray!120}{\textit{- The incorrect answer should not be obviously absurd; it must be plausible but incorrect.}} \\
      - 문제는 반드시 질문을 읽어야 풀 수 있으며, 질문이 제시되지 않으면 정답과 오답만으로는 구분이 불가능해야 합니다. \\
      \textcolor{gray!120}{\textit{- The problem must be read to solve it; without the question, the correct and incorrect answers alone should not be distinguishable.}} \\
      - 정답과 쉽게 구분되지 않도록 너무 단순하거나 쉽게 판별되는 오답은 피해야 합니다. \\
      \textcolor{gray!120}{\textit{- Avoid overly simple or easily identifiable incorrect answers that can be easily distinguished from the correct one.}} \\
      - 정답과 오답의 우열이 명백한 물리적 원리에 기반해야 합니다. \\
      \textcolor{gray!120}{\textit{- The distinction between correct and incorrect answers must be based on clear physical principles.}} \\
    - \texttt{comparison}: 정답이 왜 물리적으로 옳고, 오답은 왜 틀렸는지 명확하게 비교 설명합니다. \\
      \textcolor{gray!120}{\textit{- \texttt{comparison}: Clearly explain why the correct answer is physically valid and why the incorrect answer is wrong.}}
    \end{tcolorbox}
    \vspace{-3mm}
    \captionof{figure}{Prompt template for generating an incorrect answer. (Synthesis Stage 2)}
% \end{figure*}

\newpage

\subsection{Prompts for Verifying Questions and Answers (Verification Stage 1)}
\label{appendix:Prompts for Verifying Questions and Answers}
% \begin{figure*}[htb!]
% \vspace{3mm}
    \small
    \centering
    \begin{tcolorbox}[colback=gray!10, colframe=black, title=System Prompt for Verification Stage 1]
    당신은 AI 모델의 물리적 추론 능력을 평가하기 위한 고품질 한국어 데이터셋을 평가하는 '데이터 평가 전문가' 입니다. \\
    \textcolor{gray!120}{\textit{You are a "data evaluation expert" tasked with evaluating high-quality Korean datasets designed to assess AI models' physical reasoning abilities.}} \\
    당신의 핵심 임무는 주어지는 평가 기준과 요청에 따라 데이터를 평가하는 것입니다. \\
    \textcolor{gray!120}{\textit{Your core mission is to evaluate the data according to the given evaluation criteria and instructions.}}
    \end{tcolorbox}
    \vspace{-3mm}

    \begin{tcolorbox}[colback=gray!10, colframe=black, title=Input Prompt Template for verifying the correctness of an answer in Verification Stage 1, breakable]
    주어진 평가 기준에 따라 다음 데이터셋을 평가하십시오. \\
    \textcolor{gray!120}{\textit{Evaluate the following dataset according to the given evaluation criteria.}} \\
    
    \# 평가 기준 \\
    \textcolor{gray!120}{\textit{\# Evaluation Criteria}} \\
    - 주어진 문제에 대한 정답이 실제로 올바른 정답인지 평가하십시오. 틀린 답이라면 false, 실제로 올바른 정답이라면 true의 값으로 평가하십시오. \\
    \textcolor{gray!120}{\textit{- Assess whether the given answer to the problem is actually correct. Use \texttt{false} if it is incorrect and \texttt{true} if it is correct.}} \\
    - 평가에 대한 이유를 함께 서술하십시오. (\texttt{correctness\_rationale}) \\
    \textcolor{gray!120}{\textit{- Provide a rationale for your evaluation. (\texttt{correctness\_rationale})}} \\
    
    \# 평가 요청 \\
    \textcolor{gray!120}{\textit{\# Evaluation Request}} \\
    - JSON 객체 형식으로 평가 결과를 생성하십시오. \\
    \textcolor{gray!120}{\textit{- Generate the evaluation result in a JSON object format.}} \\
    - 평가 기준에 따라 평가한 값을 'correctness: \{\{value\}\}'로 생성하십시오. \\
    \textcolor{gray!120}{\textit{- Provide the evaluated value as 'correctness: \{\{value\}\}' according to the evaluation criteria.}} \\

    \# 데이터셋 \\
    \textcolor{gray!120}{\textit{\# Dataset}} \\

    \#\# 문제 \\
    \textcolor{gray!120}{\textit{\#\# Problem}} \\
    \{problem\} \\

    \#\# 정답 \\
    \textcolor{gray!120}{\textit{\#\# Answer}} \\
    \{answer\}
    \end{tcolorbox}

    \vspace{-3mm}
    \begin{tcolorbox}[colback=gray!10, colframe=black, title=Input Prompt Template for verifying a problem in Verification Step 1, breakable]
    주어진 평가 기준에 따라 다음 데이터셋을 평가하십시오. \\
    \textcolor{gray!120}{\textit{Evaluate the following dataset according to the given evaluation criteria.}} \\

    \# 평가 기준 \\
    \textcolor{gray!120}{\textit{\# Evaluation Criteria}} \\

    \#\# 명확성 평가 기준 \\
    \textcolor{gray!120}{\textit{\#\# Clarity Evaluation Criteria}} \\
    - 주어진 문제와 정답이 명확한지 평가하십시오. (\texttt{is\_problem\_clear}) \\
    \textcolor{gray!120}{\textit{- Evaluate whether the given problem and answer are clear. (\texttt{is\_problem\_clear})}} \\
    - 문제와 정답이 모호하거나, 해석에 혼동을 주어 의도를 파악하기 어렵다면 false로 평가하십시오. \\
    \textcolor{gray!120}{\textit{- If the problem and answer are ambiguous or confusing to interpret, evaluate as false.}} \\
    - 문제와 정답이 명확하고 혼동의 여지가 없다면 true의 값으로 평가하십시오. \\
    \textcolor{gray!120}{\textit{- If the problem and answer are clear and unambiguous, evaluate as true.}} \\

    \#\# 물리적 추론 평가 기준 \\
    \textcolor{gray!120}{\textit{\#\# Physical Reasoning Evaluation Criteria}} \\
    - 주어진 물리 요소가 문제와 정답에 포함되는지 여부를 평가하십시오. \\
    \textcolor{gray!120}{\textit{- Evaluate whether the given physical elements are included in the problem and answer.}} \\
    - 만약 주어진 물리 요소가 문제와 정답에 포함된다면 true, 포함되지 않는다면 false의 값으로 평가하십시오. (\texttt{physical\_relevance\_included}) \\
    \textcolor{gray!120}{\textit{- If the physical elements are included, evaluate as true; if not, evaluate as false. (\texttt{physical\_relevance\_included})}} \\
    \quad - 평가에 대한 이유를 함께 서술하십시오. (\texttt{physical\_relevance\_included\_rationale}) \\
    \quad \textcolor{gray!120}{\textit{- Provide the rationale for this evaluation. (\texttt{physical\_relevance\_included\_rationale})}} \\
    
    \#\# 상식 평가 기준 \\
    \textcolor{gray!120}{\textit{\#\# Commonsense Evaluation Criteria}} \\
    - 주어진 문제와 정답이 평균적인 한국인의 상식 수준으로 풀 수 있는지 true 또는 false의 값으로 평가하십시오. (\texttt{commonsense\_feasible}) \\
    \textcolor{gray!120}{\textit{- Evaluate whether the problem and answer can be solved using the commonsense level of an average Korean. (\texttt{commonsense\_feasible})}} \\
    \quad - 평가에 대한 이유를 함께 서술하십시오. (\texttt{commonsense\_feasible\_rationale}) \\
    \quad \textcolor{gray!120}{\textit{- Provide the rationale for this evaluation. (\texttt{commonsense\_feasible\_rationale})}} \\
    
    \#\# 문화 집중 평가 기준 \\
    \textcolor{gray!120}{\textit{\#\# Cultural Focus Evaluation Criteria}} \\
    - 한국적인 문화 요소란 보편적인 문화 요소를 제외한 다른 국가와는 차별적인 한국만의 특색있는 문화 요소를 의미합니다. 한국의 전통적인 요소가 될 수도 있으며, 현대적인 한국의 요소가 될 수 있습니다.\\
    \textcolor{gray!120}{\textit{- Korean cultural elements refer to uniquely Korean elements distinct from universal or common global cultural elements. They may be traditional or modern Korean elements.}}\\
    - 주어지는 문화 요소 중에서 위 한국적인 문화 요소 기준에 따라 한국적인 문화 요소에 포함되지 않는 요소를 제외하고 한국적인 문화 요소를 구성하십시오. (\texttt{korea\_relevance\_list}) \\
    \textcolor{gray!120}{\textit{- From the given cultural elements, exclude those that do not meet the uniquely Korean criteria, and construct a list of Korean cultural elements. (\texttt{korea\_relevance\_list})}} \\
    \quad - 문화 요소의 제외 또는 포함에 대한 이유를 함께 서술하십시오. (\texttt{korea\_relevance\_rationale}) \\
    \quad \textcolor{gray!120}{\textit{- Provide the rationale for inclusion or exclusion. (\texttt{korea\_relevance\_rationale})}} \\
    
    - 구성한 한국적인 문화 요소가 문제와 정답에 포함되지 않는다면 0, 한국적 문화 요소가 문제와 정답에 단순히 언급되며 다른 요소로 대체해도 문제가 성립한다면 1, 한국적 문화 요소가 문제와 정답에 핵심적으로 포함되며 한국적 문화 요소에 대한 이해가 문제 해결에 필수적이라면 2의 값을 평가하십시오. (\texttt{cultural\_element\_score}) \\
    \textcolor{gray!120}{\textit{- If the constructed Korean cultural elements are not included in the problem and answer, evaluate as 0. If the Korean cultural elements are merely mentioned in the problem and answer but can be replaced with other elements without affecting the problem, evaluate as 1. If the Korean cultural elements are essential in the problem and answer, and understanding them is crucial for solving the problem, evaluate as 2. (\texttt{cultural\_element\_score})}} \\
    \quad - 평가에 대한 이유를 함께 서술하십시오. (\texttt{cultural\_element\_score\_rationale}) \\
    \quad \textcolor{gray!120}{\textit{- Provide the rationale for this evaluation. (\texttt{cultural\_element\_score\_rationale})}} \\
    
    \#\# 유창성 평가 기준 \\
    \textcolor{gray!120}{\textit{\#\# Fluency Evaluation Criteria}} \\
    - 주어진 문제와 정답을 연결하였을 때, 한국어 문장이 자연스러운지 평가하십시오. (\texttt{fluency}) \\
    \textcolor{gray!120}{\textit{- Evaluate whether the combined problem and answer form a natural Korean sentence. (\texttt{fluency})}} \\
    - 한국어 원어민이 썼다고 느껴질 만큼 자연스럽고 유창하면 true의 값으로 평가하십시오. \\
    \textcolor{gray!120}{\textit{- If it sounds fluent and natural as if written by a native Korean, evaluate as true.}} \\
    - 번역투의 부자연스러운 표현이 포함되어 있다면 false의 값으로 평가하십시오. \\
    \textcolor{gray!120}{\textit{- If awkward, translation-like expressions are present, evaluate as false.}} \\
    
    \# 평가 요청 \\
    \textcolor{gray!120}{\textit{\# Evaluation Request}} \\
    - JSON 객체 형식으로 평가 결과를 생성하십시오. \\
    \textcolor{gray!120}{\textit{- Generate the evaluation result in a JSON object format.}} \\
    - 명확성 평가 기준에 따라 평가한 값을 'is\_problem\_clear: \{\{value\}\}'로 생성하십시오. \\
    \textcolor{gray!120}{\textit{- According to the clarity evaluation criterion, generate the value as 'is\_problem\_clear: \{\{value\}\}'.}} \\
    - 물리적 추론 평가 기준에 따라 평가한 값을 'physical\_relevance\_included: \{\{value\}\}'로 생성하십시오. \\
    \textcolor{gray!120}{\textit{- According to the physical reasoning evaluation criterion, generate the value as 'physical\_relevance\_included: \{\{value\}\}'.}} \\
    \ \ \ \ - 물리적 추론 평가 기준에 따라 평가한 이유를 'physical\_relevance\_included\_rationale: \{\{value\}\}'로 생성하십시오. \\
    \textcolor{gray!120}{\textit{- Provide the rationale for the evaluation according to the physical reasoning criterion as 'physical\_relevance\_included\_rationale: \{\{value\}\}'.}} \\
    - 상식 평가 기준에 따라 평가한 값을 'commonsense\_feasible: \{\{value\}\}'로 생성하십시오. \\
    \textcolor{gray!120}{\textit{- According to the commonsense evaluation criterion, generate the value as 'commonsense\_feasible: \{\{value\}\}'.}} \\
    \ \ \ \ - 상식 평가 기준에 따라 평가한 이유를 'commonsense\_feasible\_rationale: \{\{value\}\}'로 생성하십시오. \\
    \textcolor{gray!120}{\textit{- Provide the rationale for the evaluation according to the commonsense criterion as 'commonsense\_feasible\_rationale: \{\{value\}\}'.}} \\
    - 문화 집중 평가 기준에 따라 주어진 문제와 정답을 평가하여 'cultural\_element\_score: \{\{value\}\}'로 생성하십시오. \\
    \textcolor{gray!120}{\textit{- Evaluate the given problem and answer according to the cultural focus criterion and generate 'cultural\_element\_score: \{\{value\}\}'.}} \\
    \ \ \ \ - 문화 집중 평가 기준에 따라 평가한 이유를 'cultural\_element\_score\_rationale: \{\{value\}\}'로 생성하십시오. \\
    \textcolor{gray!120}{\textit{- Provide the rationale for the evaluation according to the cultural focus criterion as 'cultural\_element\_score\_rationale: \{\{value\}\}'.}} \\
    - 구성한 한국적인 전통요소를 'korea\_relevance\_list: \{\{list\}\}'로 생성하십시오. \\
    \textcolor{gray!120}{\textit{- Generate the composed Korean traditional elements as 'korea\_relevance\_list: \{\{list\}\}'.}} \\
    \ \ \ \ - 문화 요소의 제외 또는 포함에 대한 이유를 'korea\_relevance\_rationale: \{\{value\}\}'로 생성하십시오. \\
    \textcolor{gray!120}{\textit{- Provide the rationale for inclusion or exclusion of cultural elements as 'korea\_relevance\_rationale: \{\{value\}\}'.}} \\
    - 유창성 평가 기준에 따라 평가한 값을 'fluency: \{\{value\}\}'로 생성하십시오. \\
    \textcolor{gray!120}{\textit{- According to the fluency evaluation criterion, generate the value as 'fluency: \{\{value\}\}'.}} \\

    \# 데이터셋 \\
    \textcolor{gray!120}{\textit{\# Dataset}} \\
    
    \#\# 문제 \\
    \textcolor{gray!120}{\textit{\#\# Problem}} \\
    \{problem\} \\
    
    \#\# 정답 \\
    \textcolor{gray!120}{\textit{\#\# Answer}} \\
    \{answer\} \\
    
    \#\# 문화 요소 \\
    \textcolor{gray!120}{\textit{\#\# Cultural Elements}} \\
    \{korea\_relevance\} \\
    
    \#\# 물리 요소 \\
    \textcolor{gray!120}{\textit{\#\# Physical Elements}} \\
    \{physical\_relevance\}
    \end{tcolorbox}
    
    \vspace{-3mm}
    \captionof{figure}{Prompt template for verifying a question and an answer. (Verification Stage 1)}
% \end{figure*}

\newpage
\subsection{Prompts for Verifying Incorrect Answers (Verification Stage 2)}
\label{appendix:Prompts for Verifying Incorrect Answers}
% \begin{figure*}[htb!]
% \vspace{3mm}
    \small
    \centering
    \begin{tcolorbox}[colback=gray!10, colframe=black, title=System Prompt for Verification Stage 2]
    당신은 AI 모델의 물리적 추론 능력을 평가하기 위한 고품질 한국어 데이터셋을 평가하는 '데이터 평가 전문가' 입니다. \\
    \textcolor{gray!120}{\textit{You are a "data evaluation expert" tasked with evaluating high-quality Korean datasets designed to assess AI models' physical reasoning abilities.}} \\
    당신의 핵심 임무는 주어지는 평가 기준과 요청에 따라 데이터를 평가하는 것입니다. \\
    \textcolor{gray!120}{\textit{Your core mission is to evaluate the data according to the given evaluation criteria and instructions.}}
    \end{tcolorbox}
    \vspace{-3.5mm}
    
    \begin{tcolorbox}[colback=gray!10, colframe=black, title=Input Prompt Template for Verification Stage 2, breakable]
    주어진 평가 기준에 따라 다음 데이터셋을 평가하십시오. \\
    \textcolor{gray!120}{\textit{Evaluate the following dataset according to the given evaluation criteria.}} \\
    
    \# 평가 기준 \\
    \textcolor{gray!120}{\textit{\# Evaluation Criteria}} \\
    
    \#\# 오답 평가 기준 \\
    \textcolor{gray!120}{\textit{\#\# Incorrect Answer Evaluation Criteria}} \\
    - 주어진 오답이 주어진 문제에 대해 실제로 틀린 답인지 평가하십시오. (is\_true\_distractor) \\
    \textcolor{gray!120}{\textit{- Evaluate whether the given incorrect answer is truly wrong for the given problem. (is\_true\_distractor)}} \\
    - 만약 주어진 오답이 실제로 문제에 대한 정답이 아니라면 (즉 오답이라면) 오답에 대한 기능을 제대로 수행하고 있으므로 true의 값으로 평가하십시오. \\
    \textcolor{gray!120}{\textit{- If the given incorrect answer is not actually the correct answer to the problem (i.e., it is indeed incorrect), then it serves its function properly and should be evaluated as true.}} \\
    - 만약 주어진 오답이 실제로 문제에 대한 정답으로 볼 수 있거나, 다소 모호한 정답이라면 오답에 대한 기능을 제대로 수행하지 못하고 있으므로 false의 값으로 평가하십시오. \\
    \textcolor{gray!120}{\textit{- If the given incorrect answer can actually be seen as a correct or somewhat ambiguous answer, then it does not serve its function as an incorrect answer and should be evaluated as false.}} \\
    
    \#\# 오답 완전성 평가 기준 \\
    \textcolor{gray!120}{\textit{\#\# Completeness Evaluation Criteria for Incorrect Answers}} \\
    - 오답 선택지 자체에 모순이나 불완전한 부분이 없는지 평가하십시오. (is\_distractor\_complete) \\
    \textcolor{gray!120}{\textit{- Evaluate whether the incorrect answer option itself has no contradictions or incompleteness. (is\_distractor\_complete)}} \\
    - 만약 오답 선택지 자체에 모순이나 불완전한 부분이 없다면 true의 값으로 평가하십시오. \\
    \textcolor{gray!120}{\textit{- If the incorrect answer option has no contradictions or incompleteness, evaluate it as true.}} \\
    - 만약 오답 선택지 자체에 모순이나 불완전한 부분이 있다면 false의 값으로 평가하십시오. \\
    \textcolor{gray!120}{\textit{- If the incorrect answer option has contradictions or incompleteness, evaluate it as false.}} \\
    
    \#\# 오답 유형 평가 기준 \\
    \textcolor{gray!120}{\textit{\#\# Type Evaluation Criteria for Incorrect Answers}} \\
    - 정답과 오답의 우열이 명백한 물리적 원리에 기반하는지 평가하십시오. (\texttt{is\_phyiscally\_distinct}) \\
    \textcolor{gray!120}{\textit{- Evaluate whether the superiority of the correct answer over the incorrect answer is clearly based on physical principles. (\texttt{is\_phyiscally\_distinct})}} \\
    - 만약 정답과 오답의 우열이 명백한 물리적 원리에 기반한다면 true의 값으로 평가하십시오. \\
    \textcolor{gray!120}{\textit{- If the superiority of the correct answer over the incorrect answer is clearly based on physical principles, evaluate it as true.}} \\
    - 만약 정답과 오답의 우열이 명백한 물리적 원리에 기반하지 않고 주관적인 선호, 비물리적인 요인등에 의한 것이라면 false의 값으로 평가하십시오. \\
    \textcolor{gray!120}{\textit{- If the superiority is not based on physical principles but rather on subjective preferences or non-physical factors, evaluate it as false.}} \\
    
    \# 평가 요청 \\
    \textcolor{gray!120}{\textit{\# Evaluation Request}} \\
    - JSON 객체 형식으로 평가 결과를 생성하십시오. \\
    \textcolor{gray!120}{\textit{- Generate the evaluation result in JSON object format.}} \\
    - 오답 평가 기준에 따라 평가한 값을 'is\_true\_distractor: \{\{value\}\}'로 생성하십시오. \\
    \textcolor{gray!120}{\textit{- According to the incorrect answer evaluation criterion, generate the value as 'is\_true\_distractor: \{\{value\}\}'.}} \\
    - 오답 완전성 평가 기준에 따라 평가한 값을 'is\_distractor\_complete: \{\{value\}\}'로 생성하십시오. \\
    \textcolor{gray!120}{\textit{- According to the completeness evaluation criterion for incorrect answers, generate the value as 'is\_distractor\_complete: \{\{value\}\}'.}} \\
    - 오답 유형 평가 기준에 따라 평가한 값을 'is\_phyiscally\_distinct: \{\{value\}\}'로 생성하십시오. \\
    \textcolor{gray!120}{\textit{- According to the type evaluation criterion for incorrect answers, generate the value as 'is\_phyiscally\_distinct: \{\{value\}\}'.}} \\
    
    \# 데이터셋 \\
    \textcolor{gray!120}{\textit{\# Dataset}} \\

    \#\# 문제 \\
    \textcolor{gray!120}{\textit{\#\# Problem}} \\
    \{problem\} \\
    
    \#\# 정답 \\
    \textcolor{gray!120}{\textit{\#\# Answer}} \\
    \{answer\} \\
    
    \#\# 오답 \\
    \textcolor{gray!120}{\textit{\#\# Incorrect Answer}} \\
    \{incorrect\_answer\}
    \end{tcolorbox}
    \vspace{-3mm}
    \captionof{figure}{Prompt template for verifying an incorrect answer. (Verification Stage 2)}
% \end{figure*}

\newpage

\subsection{Prompts for Filtering Answer Bias}
\label{appendix:Prompts for Filtering Answer Bias}
% \begin{figure*}[htb!]
% \vspace{3mm}
    \small
    \centering
    \begin{tcolorbox}[colback=gray!10, colframe=black, title=System Prompt for Filtering Answer Bias]
    당신은 문제를 보지 않고, 주어진 선지만으로 정답을 찾을 수 있는지 판단해야 합니다. \\
    \textcolor{gray!120}{\textit{You need to determine whether you can find the correct answer based solely on the given choices without seeing the question.}} \\
    주의사항: \\
    \textcolor{gray!120}{\textit{Instructions:}} \\
    - 질문은 제공되지 않았습니다. 단지 두 선지만 보고 정답을 고르세요. \\
    \textcolor{gray!120}{\textit{- The question is not provided. Choose the correct answer by looking only at the two options.}} \\
    - 선지의 길이나 특정 단어 패턴 등 표면적 특징만 보고 판단하지 마세요. \\
    \textcolor{gray!120}{\textit{- Do not make your judgment based only on superficial features such as the length of the choices or specific word patterns.}} \\
    - 정답 번호만 출력하세요. \\
    \textcolor{gray!120}{\textit{- Output only the number of the correct answer.}} \\
    - 두 선지만으로 정답을 판단할 수 없다면 3을 출력하세요. \\
    \textcolor{gray!120}{\textit{- If you cannot determine the correct answer from the two choices alone, output 3.}} \\
    - 이유나 설명은 출력하지 마세요. \\
    \textcolor{gray!120}{\textit{- Do not output any reasons or explanations.}}
    \end{tcolorbox}
    \vspace{-3mm}
    
    \begin{tcolorbox}[colback=gray!10, colframe=black, title=Input Prompt Template for Filtering Answer Bias, breakable]
    다음 중 정답을 고르세요. \\
    \textcolor{gray!120}{\textit{Select the correct answer from the following options.}} \\
    1. \{solution0\} \\
    2. \{solution1\} \\
    3. 두 선지만으로 정답을 판단할 수 없다. \\
    \textcolor{gray!120}{\textit{3. You cannot determine the correct answer from the two choices alone.}} \\
    정답 번호만 출력하세요. 1 또는 2 또는 3 중 하나만 선택합니다. \\
    \textcolor{gray!120}{\textit{Output only the number of the correct answer. Choose only one among 1, 2, or 3.}}
    \end{tcolorbox}
    \vspace{-5mm}
    \captionof{figure}{Prompt template for filtering answer bias.}
% \end{figure*}
\vspace{3mm}

\subsection{Prompts for Evaluation}
\label{appendix:Prompts for Evaluation}
% \begin{figure*}[htb!]
% \vspace{3mm}
    \small
    \centering
    \begin{tcolorbox}[colback=gray!10, colframe=black, title=System Prompt for Evaluation]
    당신은 주어진 문제에 대해 제시된 두 개의 선택지 중 적절한 정답 하나를 고르는 AI 어시스턴트입니다. \\
    \textcolor{gray!120}{\textit{You are an AI assistant who selects the correct answer from two given options for a specific problem.}} \\
    주의사항: \\
    \textcolor{gray!120}{\textit{Instructions:}} \\
    - 정답 번호만 출력하세요. 반드시 1 또는 2 중 하나만 선택합니다. \\
    \textcolor{gray!120}{\textit{- Output only the number of the correct answer. You must select either 1 or 2.}} \\
    - 이유나 설명은 출력하지 마세요. \\
    \textcolor{gray!120}{\textit{- Do not output any reasons or explanations.}}
    \end{tcolorbox}
    \vspace{-3mm}
    
    \begin{tcolorbox}[colback=gray!10, colframe=black, title=Input Prompt Template for Evaluation, breakable]
    질문: \{question\} \\
    \textcolor{gray!120}{\textit{Question: \{question\}}} \\
    1. \{solution0\} \\
    2. \{solution1\} \\
    정답 번호만 입력하세요. 1 또는 2 \\
    \textcolor{gray!120}{\textit{Input only the correct answer number. 1 or 2}}
    \end{tcolorbox}
    \vspace{-5mm}
    \captionof{figure}{Prompt template for evaluation.}
% \end{figure*}

\newpage
\section{Human Evaluation Guide}
\label{appendix:Human Evaluation Guide}

\begin{flushleft}
To ensure a standardized and rigorous final review, we provided the human expert with a comprehensive evaluation guide. This guide outlines 5 key assessment criteria: accuracy, naturalness and logical consistency, cultural appropriateness, common sense, and plausibility. The complete set of instructions is available in Figure \ref{fig:eval_guidelines}, which details the standards for accepting or rejecting a data instance.
\end{flushleft}

\begin{tcolorbox}[colback=gray!10, colframe=black, title=Human Evaluation Guide, breakable]
    \textbf{1. 평가 개요} \\
    \textcolor{gray!120}{\textbf{1. Evaluation Overview}} \\
    이 문서는 \texttt{dataset.jsonl} 파일에 포함된 한국어 질의응답 데이터의 품질을 평가하기 위한 가이드입니다. \\
    \textcolor{gray!120}{\textit{This document is a guide for evaluating the quality of Korean Q\&A data included in the \texttt{dataset.jsonl} file.}} \\
    각 항목은 다음 요소들을 포함하고 있습니다: \\
    \textcolor{gray!120}{\textit{Each item includes the following:}} \\
    - 문제 유형 카테고리 \\
    \textcolor{gray!120}{\textit{- Question Type Category}} \\
    - 문제 유형 서브카테고리 \\
    \textcolor{gray!120}{\textit{- Question Type Subcategory}} \\
    - 프롬프트 (질문) \\
    \textcolor{gray!120}{\textit{- Prompt (Question)}} \\
    - 두 개의 답변 (\texttt{solution0}, \texttt{solution1}) \\
    \textcolor{gray!120}{\textit{- Two answers (\texttt{solution0}, \texttt{solution1})}} \\
    - 정답 레이블 \\
    \textcolor{gray!120}{\textit{- Correct answer label}} \\
    - 메타데이터 (문화적 관련성, 물리적 관련성 점수 등) \\
    \textcolor{gray!120}{\textit{- Metadata (cultural relevance, physical relevance scores, etc.)}} \\
    \vspace{3mm}

    \textbf{2. 평가 항목} \\
    \textcolor{gray!120}{\textit{\textbf{2. Evaluation Items}}} \\
    \textbf{2.1 정확성 (Accuracy)} \\
    \textcolor{gray!120}{\textit{\textbf{2.1 Accuracy}}} \\
    - 제시된 질문에 따라 두 답변 중 정답을 구하세요. \\
    \textcolor{gray!120}{\textit{- Find the correct answer out of the two solutions provided based on the question.}} \\
    \vspace{2mm}

    \textbf{2.2 자연스러움과 논리성 (Naturalness \& Logical Consistency)} \\
    \textcolor{gray!120}{\textit{\textbf{2.2 Naturalness \& Logical Consistency}}} \\
    - 질문과 답변이 자연스럽고 올바른 한국어로 작성되었는지 확인하세요. \\
    \textcolor{gray!120}{\textit{- Check if the question and answer are written in natural and correct Korean.}} \\
    - 어색한 표현이나 문법 오류가 없는지 점검하세요. \\
    \textcolor{gray!120}{\textit{- Check for awkward expressions or grammatical errors.}} \\
    - 일상생활에서 실제로 사용할 법한 표현인지 고려하세요. \\
    \textcolor{gray!120}{\textit{- Consider whether the expressions are commonly used in daily life.}} \\
    - 질문과 답변 사이에 논리적 일관성이 유지되는지 확인하세요. \\
    \textcolor{gray!120}{\textit{- Check if logical consistency is maintained between the question and answer.}}
    \vspace{2mm}

    \textbf{2.3 문화적 적절성 (Cultural Appropriateness)} \\
    \textcolor{gray!120}{\textit{\textbf{2.3 Cultural Appropriateness}}} \\
    - 한국 문화와 관련된 내용이 정확하게 반영되었는지 평가하세요. \\
    \textcolor{gray!120}{\textit{- Evaluate whether content related to Korean culture is accurately reflected.}} \\
    - 전통 문화에 대한 잘못된 정보가 없는지 확인하세요. \\
    \textcolor{gray!120}{\textit{- Check for any misinformation about traditional culture.}} \\
    - 문화적 민감성을 고려하여 부적절한 표현이 없는지 검토하세요. \\
    \textcolor{gray!120}{\textit{- Review for inappropriate expressions, considering cultural sensitivity.}} \\
    \vspace{2mm}

    \textbf{2.4 상식 (Common Sense)} \\
    \textcolor{gray!120}{\textit{\textbf{2.4 Common Sense}}} \\
    - 문제가 한국인의 상식 수준에서 풀 수 있는지 확인하세요. \\
    \textcolor{gray!120}{\textit{- Check if the problem can be solved at the level of a Korean person's common sense.}}
    \vspace{2mm}

    \textbf{2.5 현실성 (Practicality / Plausibility)} \\
    \textcolor{gray!120}{\textit{\textbf{2.5 Practicality / Plausibility}}} \\
    - 문제와 답변이 실제 상황에서 합리적이고 실행 가능해야 합니다. \\
    \textcolor{gray!120}{\textit{- The question and answer must be reasonable and plausible in a real-world situation.}} \\
    - 물리적, 사회적, 상식적으로 불가능하거나 극도로 비현실적인 선택지는 배제해야 합니다. \\
    \textcolor{gray!120}{\textit{- Exclude options that are physically, socially, or common-sensically impossible or extremely unrealistic.}} \\
    \vspace{3mm}

    \textbf{3. 평가 방법} \\
    \textcolor{gray!120}{\textit{\textbf{3. Evaluation Method}}} \\
    1. 각 항목을 주의 깊게 읽고 이해하세요. \\
    \textcolor{gray!120}{\textit{1. Read and understand each item carefully.}} \\
    2. 제공된 정답 레이블이 올바른지 확인하세요. \\
    \textcolor{gray!120}{\textit{2. Check if the provided correct answer label is correct.}} \\
    3. 각 평가 항목(정확성, 자연스러움과 논리성, 문화적 적절성, 상식, 현실성)에 대해 0,1점으로 평가하세요. \\
    \textcolor{gray!120}{\textit{3. Evaluate each item (Accuracy, Naturalness \& Logical Consistency, Cultural Appropriateness, Common Sense, Practicality) with a score of 0 or 1.}}
    \vspace{3mm}

    \textbf{4. 주의사항} \\
    \textcolor{gray!120}{\textit{\textbf{4. Precautions}}} \\
    - 모든 평가는 객관적으로 진행해 주세요. \\
    \textcolor{gray!120}{\textit{- Please conduct all evaluations objectively.}}

\end{tcolorbox}
\vspace{-5mm}
\captionof{figure}{Guide for human evaluation.}
\label{fig:eval_guidelines}

\end{document}